\documentclass[11pt]{article}
\usepackage[a4paper, total={6.5in, 9in}]{geometry}
\usepackage{latexsym}
\usepackage{amsmath}

\usepackage{amssymb}
\usepackage{amsthm}
\usepackage{float}
\usepackage{mathtools}
\usepackage{xcolor}
\usepackage{hyperref}
\usepackage{bbm}
\usepackage{comment}
\usepackage{algorithm}
\usepackage{algpseudocode}
\usepackage{epsfig}
\usepackage[shortlabels]{enumitem}
\usepackage{mathtools, nccmath}
\usepackage[square,numbers]{natbib}
\DeclareMathOperator{\prob}{\mathbb{P}}
\DeclareMathOperator{\argmax}{argmax}

\newcommand{\calS}{{\mathcal{S}}}
\newcommand{\calA}{{\mathcal{A}}}
\newcommand{\calD}{{\mathcal{D}}}
\newcommand{{\calY}}{{\mathcal{Y}}}

\newcommand{\calG}{{\mathcal{G}}}

\newcommand{\calR}{{\mathcal{R}}}
\newcommand{\calT}{{\mathcal{T}}}
\newcommand{\calO}{{\mathcal{O}}}
\newcommand{\calP}{{\mathcal{P}}}

\newcommand{\bary}{\Bar{{y}}}

\newcommand{\barprob}{{\Bar{\prob}}}
\newcommand{\barr}{{\Bar{{r}}}}
\newcommand{\barR}{{\Bar{R}}}
\newcommand{\barV}{{\Bar{V}}}
\newcommand{\barQ}{{\Bar{Q}}}
\newcommand{{\barpi}}{{\Bar{\pi}}}
\newcommand{\barPi}{{\Bar{\Pi}}}
\newcommand{\barY}{{\Bar{\calY}}}
\newcommand{\barI}{{\Bar{I}}}
\newcommand{\hatV}{{\Hat{V}}}
\newcommand{\hatQ}{{\Hat{Q}}}
\newcommand{\hatpi}{{\Hat{\pi}}}
\newcommand{\hatprob}{{\Hat{\prob}}}
\newcommand{\hatR}{{\Hat{R}}}

\newcommand{\R}{\mathbb{R}}
\newcommand{\E}{\mathbb{E}}

\newcommand{\MDP}{\texttt{RS-MDP}(\mathcal{S},\mathcal{A},H,\prob^{S},R,U)}
\newcommand{\RSMDP}{\texttt{RS-MDP}(\calS, \calA, H, \prob^S,R,U)}
\newcommand{\BARMDP}{{\texttt{RS-MDP}}(\mathcal{S},\mathcal{A},H,\prob^S,\barR,U)}
\newcommand{\HATMDP}{{\texttt{RS-MDP}}(\mathcal{S},\mathcal{A},H,\hat{\prob}^S,\hat{R},U)}
\newcommand{\projY}[1]{{\rm Proj}_{\barY_h}(#1)}
\newcommand{\projR}[1]{{\rm Proj}_{\barI}(#1)}

\newcommand{\classRSMAB}{{\Xi}_{\rm RS-MAB}}
\newcommand{\classRSMDP}{{\Xi}_{\rm RS-MDP}}
\newcommand{\algdetclass}{{\rm ALG^{\rm det}}}
\newcommand{\algr}{{\rm alg}^{\rm rand}}
\newcommand{\algd}{{\rm alg}^{\rm det}}
\newcommand{\algrandclass}{{\mathcal{P}({\rm ALG^{\rm det}})}}
\hypersetup{
    colorlinks=true,
    linkcolor=blue,
    filecolor=magenta,      
    urlcolor=blue,
    }
\everymath=\expandafter{\the\everymath\displaystyle}

\newcommand{\infnorm}[1]{\left\| #1\right\|_{\infty}}
\newtheorem{theorem}{Theorem}
\newtheorem{remark}{Remark}
\newtheorem{corollary}{Corollary}
\newtheorem{lemma}{Lemma}

\newtheorem{proposition}{Proposition}
\newtheorem{definition}{Definition}

\newtheorem{assumption}{Assumption}

\DeclareMathSizes{24}{24}{24}{24}
\title{Risk-sensitive Markov Decision Process and Learning under General Utility Functions}

\author{Zhengqi Wu
\thanks{Daniel J. Epstein Department of Industrial and Systems Engineering, University of Southern California, Los Angeles, CA 90089, USA. \textbf{Email:} zhengqiw@usc.edu}
\and
Renyuan Xu 
\thanks{Department of Finance and Risk Engineering, New York University, New York, NY 10012, USA. \textbf{Email:} rx2364@nyu.edu. R.X. is partially supported
by the NSF CAREER award DMS-2339240 and a JP Morgan Faculty Research Award.}
}
\date{First version: October 2023\\
This version: \today}
\begin{document}
\maketitle
\begin{abstract}
Reinforcement Learning (RL) has gained substantial attention across diverse application domains and theoretical investigations. Existing literature on RL theory largely focuses on risk-neutral settings where the decision-maker learns to maximize the expected cumulative reward. 
However, in practical scenarios such as portfolio management and e-commerce recommendations, clients often exhibit heterogeneous risk preferences. While incorporating these preferences can be approached through utility theory, the development of risk-sensitive RL under general utility functions remains an open question for theoretical exploration.

In this paper, we consider a scenario where the decision-maker aims to optimize a general utility function of the cumulative reward in the framework of a Markov decision process (MDP) in an unknown environment. To facilitate the Dynamic Programming Principle and Bellman equation, we enlarge the state space with an additional dimension that accounts for the cumulative reward. We propose a discretized approximation scheme for the MDP under the enlarged state space, which is computationally tractable and crucial for algorithmic design. Building on this, we propose a modified value iteration algorithm that employs an epsilon-covering of the space of cumulative reward. In the case with access to a simulator,  we demonstrate that our algorithm efficiently identifies a near-optimal policy, achieving a sample complexity of $\Tilde{\calO}\left(\frac{H^7SA\kappa^2 }{\epsilon^2}\right)$, where $S$ represents the number of states, $A$ the number of actions, $H$ the length of an episode, and $\kappa$ the Lipschitz coefficient of the utility function. In the case without an accessible simulator, we develop an algorithm that leverages upper-confidence-bound exploration, achieving a regret bound of {$\Tilde{\calO}\left(\sqrt{H^4S^2AT\kappa(\lambda+\eta)}+\sqrt{H^3SAT\kappa^2}\right)$}\footnote{The term $\sqrt{H^4S^2AT\kappa(\lambda+\eta)}$ captures the leading-order dependency on $H$, $S$, $A$, and $T$, whereas the term $\sqrt{H^3SAT\kappa^2}$ reflects the leading-order dependency on the Lipschitz constant of the utility function, $\kappa$.}, where $T$ represents the total number of time steps, and $\lambda$ and $\eta$ are the regularity constants of the reward distribution. {Finally, we establish a novel theoretical lower bound for the risk-sensitive setting, and show that the regret of our algorithm matches this lower bound up to a small polynomial factor in $H$ and $S$.}
\end{abstract}
\section{Introduction}
Data-driven decision-making, particularly in the field of Reinforcement Learning (RL), has gained  substantial attention across diverse domains, including robotics control \citep{levine2016end,lillicrap2015continuous}, AlphaGo and Atari games \citep{mnih2013playing,silver2016mastering}, autonomous driving \citep{li2019reinforcement},
and algorithmic trading \citep{hambly2023recent}. 
RL involves learning via trial and error, through interactions with an unknown environment often characterized as a Markov decision process (MDP). In the framework of RL, an agent takes actions and receives reinforcement signals in terms of numerical rewards encoding the outcome of the chosen action. In the standard risk-neutral setting, the agent learns to select actions by leveraging past experiences (exploitation) and making new
choices (exploration)  to maximize the {\it expected cumulative reward} over time. Despite the extensive theoretical investigations in RL as referenced in \citep{agarwal2020model,azar2017minimax,jin2018q,jin2020provably,jin2021pessimism,wang2020reinforcement,yang2020reinforcement}, the existing literature primarily focuses on risk-neutral settings. This focus tends to overlook critical factors such as the decision-maker's attitude towards outcome uncertainty and risk preference.

In practice, many applications involve decision-makers with heterogeneous risk preferences subject to {\color{black}reward} outcome uncertainties (or stochasticity), not adequately addressed by the risk-neutral framework. Examples of these diverse risk preferences are evident in various domains, including product recommendations and pricing strategies on e-commerce platforms \citep{huang2011designing,lopez2008customer}, portfolio management and hedging in financial markets \citep{duffie1997hedging,he2011portfolio,zhou2000continuous}, and treatment decisions and resource allocations in healthcare systems \citep{johnson2021precision,valdez2019users}.
From the perspective of the e-commerce market, comprehending consumers' browsing and purchasing behavior is crucial for platforms, as it unlocks the potential to boost revenue and provide outstanding services. In contrast to shopping at retail stores, consumers on e-commerce platforms often face the challenge of not being able to physically touch or try products before making a purchase. As a result, they must make decisions while being {\it uncertain} about the potential {\color{black}reward} outcomes, such as the satisfaction they will derive from the product. 
To investigate individuals' {\it risk preferences} when facing choices that have uncertain outcomes, economists have introduced the hypothesis of Expected Utility (EU) {\color{black}\citep{von2007theory,bernardo1979expected,Mongin1998-MONEUT,schoemaker1982expected}}.
This hypothesis models the subjective value attributed to an individual's choice as the statistical expectations of their valuations regarding the outcomes of that particular choice, where the valuation is commonly modeled by a {\it utility function} satisfying certain properties.

To incorporate risk preferences in sequential decision-making and learning, we consider a scenario where the decision-maker seeks to optimize a general utility function of the cumulative reward under the framework of an MDP. 
{\color{black} }
By employing the general utility function in the objective function, the optimal policy may no longer be Markovian and may become history-dependent.
As a consequence, powerful techniques such as the Dynamic Programming Principle (DPP) and the Bellman equation become inapplicable. This renders most existing RL algorithms built upon these techniques insufficient for tackling such scenarios.

\paragraph{Our Work and Contributions.} To facilitate the application of the DPP and the Bellman equation, we enlarge the state space with an additional dimension that accounts for the cumulative reward. We prove that the value function of the enlarged problem, defined on the enlarged state space, satisfies the Bellman equation and there exists an optimal policy that is Markovian. In addition, we propose a discretized approximation scheme for the MDP under the enlarged state space, which is tractable and key for the algorithmic design. {\color{black}For this part, our key contribution is the careful analysis the value function, which allows us to efficiently bound the errors introduced by discretization. This analysis is non-trivial because even if the utility function is Lipschitz continuous, the value function of a general Markovian policy may not be Lipschitz continuous unless special structural assumptions are further specified. In our approach, 
we demonstrate that the value function of a near-optimal policy exhibits a ``near-Lipschitz'' continuity property with respect to the cumulative reward (see Proposition \ref{lemma:Lipschitz_Vbar_pihat}). Additionally, we develop a crucial result showing that regret in the discretized MDP environment can be translated into regret in the original MDP context (see Lemma \ref{lemma_regret_conversion}).  These theoretical results lay out the mathematical foundation of our RL algorithms, which could learn a near-optimal policy with low sample complexity or low regret.}

In the context of learning, we present two algorithms designed for situations both with and without simulators. Given a simulator of the environment, we propose a modified value iteration algorithm, denoted as  Value Iteration under General Utility (VIGU), that employs an epsilon-covering of the additional dimension. Under mild assumptions on the utility function {\color{black}(strictly increasing and Lipschitz continuous)}, we prove that our algorithm can efficiently identify a near-optimal policy with an $\Tilde{\calO}\left(\frac{H^7SA\kappa^2 }{\epsilon^2}\right)$ sample complexity, where $S$ represents the number of states, $A$ the number of actions, $H$ the length of an episode, and $\kappa$ the Lipschitz coefficient of the utility function.  Without access to a simulator, we propose an upper-confidence-bound (UCB) based algorithm, denoted as Value Iteration under General Utility with Upper Confidence Bound (VIGU-UCB). We prove that the proposed algorithm enjoys a regret bound on the order of {$\Tilde{\calO}\left(\sqrt{H^4S^2AT\kappa(\lambda+\eta)}+\sqrt{H^3SAT\kappa^2}\right)$}\footnote{The term $\sqrt{H^4S^2AT\kappa(\lambda+\eta)}$ captures the leading-order dependency on $H$, $S$, $A$, and $T$, whereas the term $\sqrt{H^3SAT\kappa^2}$ reflects the leading-order dependency on the Lipschitz constant of the utility function, $\kappa$. The complete expression for the regret upper bound is formally presented in Theorem \ref{thm:VIUCB_regret}.}, where $T$ represents the total number of time steps, and $\lambda$ and $\eta$ are the regularity constants of the reward distribution. To the best of our knowledge, this is the first paper providing sample complexity and regret analysis for RL under general utility functions. In the simulator setting, our sample complexity of the VIGU Algorithm matches the theoretical lower bound  ${\Omega}\left(\frac{H^4SA\kappa^2 }{\epsilon^2}\right)$ for the risk-neutral setting \citep{azar2017minimax,agarwal2020model}, up to a cubic factor in $H$ and a Lipschitz constant, and a logarithmic factor of other model parameters.  In the episodic setting without an accessible simulator, we prove that the regret of the VIGU-UCB Algorithm matches the lower bound $\Omega\left(\sqrt{H^2SAT}\right)$ for the risk-neutral setting \citep{azar2017minimax,agarwal2020model}, up to a polynomial term in  $H$, $S$, some Lipschitz constants, and a logarithmic factor of other model parameters.

{\color{black}
In addition, we establish a theoretical regret lower bound for the RL under general utility functions setting, of order $\Omega\Big(\sqrt{H S A T\underline{\kappa}^2}\Big)$, where $\underline{\kappa}>0$ denotes the lower bound on the gradient of the utility function. Previous studies have explored regret lower bounds for risk-sensitive RL, but only in very restrictive settings with limited results.  In particular, \citet{fei2020risk} and \citet{liang2024bridging} established regret lower bounds for RL under an exponential utility objective, while \citet{wang2023near} provided lower bounds for risk-sensitive MAB and RL with a CVaR objective. 
In this work, we present lower bounds for risk-sensitive RL with a general utility objective, which, to the best of our knowledge, are the first for this type of problem. This lower bound strengthens our theoretical contributions by demonstrating that the regret of our VI-UCB algorithm matches the lower bound in terms of $A$ and $T$, with only a small polynomial gap in $H$, $S$, and an additional (quadratic) dependence on the Lipschitz constant  $\kappa$. This gap arises from discretizing the enlarged state dimension for cumulative rewards, opening the door for interesting future directions to explore.
}


\paragraph{Related Work.} Following the seminal work of \citet{howard1972risk}, there have been studies on risk-sensitive MDPs (when the model is known) and, more recently, on risk-sensitive RL (when the model is unknown), exploring these topics from various perspectives. The idea of state enlargement (or state augmentation) for solving risk-sensitive decision-making problems traces back to \citet{kreps1977decision}, and was later explored by \citet{bauerle2014more,haskell2015convex,isohatala2023dynamic}. Specifically, \citet{kreps1977decision} proposed the framework of state enlargement and provided optimality conditions for history-dependent policies. \citet{bauerle2014more} established the DPP and examined the optimal Markovian policies on the enlarged state space with deterministic reward functions. They also proved the asymptotic convergence of both value-iteration-based and policy-iteration-based methods for computing the optimal value function (when the model is known). \citet{haskell2015convex} formulated a risk-sensitive control problem under general risk functionals. They showed that the problem can be formulated as an infinite-dimensional linear program (LP) in occupation measures on the enlarged state space. More recently, \citet{isohatala2023dynamic} established the theoretical framework for a continuous-time setting using a martingale approach.

Among various utility functions, the exponential function is widely adopted for its tractability and has been extensively explored in the literature. A significant amount of research on the exponential utility function \citep{borkar2002q,cavazos2011discounted,hernandez1996risk,howard1972risk,huang2020stochastic,james2004risk,mihatsch2002risk,osogami2012robustness,patek2001terminating,shen2013risk} has been conducted within the framework of MDPs under complete model information, or by establishing asymptotic convergence in the RL regime. However, the finite sample complexity has not been addressed in this line of work. Additionally, some recent works \citep{fei2020risk,fei2021exponential} proposed value iteration and Q-learning algorithms for exponential utilities with a regret of order  $\Tilde{\calO}\Big(\frac{e^{3|\beta|H}-1}{|\beta| }\sqrt{SAT}\Big)$ where $\beta$ is the risk-aversion coefficient. To the best of our knowledge, no existing theoretical results address RL under (more) general utility functions.

Another related area of study in risk-sensitive RL focuses on dynamic Markov coherent risk measures,  a class of risk measures that are coherent and time-consistent. 
Under this framework,
\citet{ruszczynski2010risk} proposed a policy iteration method and proved its asymptotic convergence guarantee. \citet{shen2014risk}
proposed a Q-learning method with an asymptotic convergence guarantee.   \citet{coache2023reinforcement} explored RL in the context of convex risk measures and developed algorithms based on policy gradients to learn the optimal policy. 
Moreover, \citet{huang2021convergence} designed a policy gradient method for MDP with Markov coherent risk and proved its convergence toward the stationary point.
Recently,  \citet{coache2022conditionally} examined dynamic spectral risk, a subset of dynamic coherent risk measures. They found that dynamic spectral risk measures can be well-approximated by neural networks and proposed an actor-critic-based RL algorithm based on this finding. {\citet{xu2023regret} studied the case where the objective is defined recursively using the optimized certainty equivalent, a family of risk measures. They designed a value-iteration-based algorithm with a regret guarantee. }

As a popular coherent risk measure that helps in understanding the losses in the worst-case scenarios, conditional value at risk (CVaR)  has been incorporated into RL in some very recent works \citep{bastani2022regret,wang2023near}. Specifically, \citet{bastani2022regret} proposed a value-iteration-based algorithm achieving an $\Tilde{\calO}(\tau^{-1}\sqrt{HSAK})$ regret where $\tau$ is the risk tolerance of CVaR. \citet{wang2023near} improved this result up to polynomial factors. However, algorithms based on risk measures often rely on nested Monte Carlo simulations, making them computationally intractable in many scenarios.

Finally, it is worth mentioning that the case of risk-neutral RL  can be recovered from our setting by choosing a linear utility function. Under the risk-neutral setting, \citet{azar2017minimax} proved that the lower bound of the regret is of order $\Omega(\sqrt{H^2SAT})$, and also showed that the value iteration method with upper confidence bound can achieve this optimal regret.  \\

The paper is organized as follows. Section \ref{sec:setting} presents the setup of risk-sensitive MDP and outlines key mathematical properties of the framework. With an accessible simulator, Section \ref{sec:VIGU} details the VIGU Algorithm along with its sample complexity analysis. For scenarios where a simulator is not available, Section \ref{sec:VIGU-UCB_simulator} introduces the VIGU-UCB Algorithm and its regret analysis. {\color{black}Lastly, Section \ref{sec:regret_LB} establishes a novel theoretical regret lower bound.}

\section{Problem Set-up and Preliminary Analysis}
\label{sec:setting}
{\color{black} Let us start with the problem set-up and some mathematical properties. Specifically, we introduce the episodic risk-sensitive MDP framework and outline its key components in Section \ref{sec:setting_RSMDP}. We then define an enlarged state space to establish DPP in Section \ref{sec:enlarged_state_space}, showing that an optimal Markovian policy exists. To address the challenge of learning in an (enlarged) continuous space, we establish key properties such as the Lipschitz continuity of the optimal value function, enabling a discretized approximation of the risk-sensitive MDP. Finally, in Section \ref{sec:discretized_env}, we bound the difference in the value functions between the original and discretized environments, which guide the development of efficient algorithms.}

\subsection{Risk-sensitive Episodic Markov Decision Process}
\label{sec:setting_RSMDP}
Let us begin by introducing the framework of episodic risk-sensitive Markov decision processes (RS-MDPs), denoted as $\MDP$, following the works of  \citet{kreps1977decision} and \citet{bauerle2014more}.  In this framework, $\mathcal{S}$ denotes the {\color{black}finite} state space,  $\mathcal{A}$ the finite action space, and {$H$}   the length of episodes. Additionally, let $S:=|\calS|$ and $A:=|\calA|$  denote the cardinalities of the state and action spaces, respectively. $\prob^S = \{\prob^S_h\}_{h\in [H]}$ is a collection of transition kernels at each timestamp $h\in [H]$, with $\prob^S_h: \mathcal{S}\times \mathcal{A} \rightarrow \calP(\mathcal{S})$ and $ [H] = \{1,2,\cdots,H\}$. 
Finally, $R = \{R_h\}_{h\in [H]}$ denotes a collection of reward distributions with $R_h: \mathcal{S}\times \mathcal{A} \rightarrow \calP([0,1])$ that has a density function.   Here  $\mathcal{P}(\mathcal{X} )$ denotes the space of probability measures on space $\mathcal{X}$.

At timestamp $h \in [H]$, the decision-maker (or the agent) takes an action $a_h \in \mathcal{A}$  given the current state $s_h \in \mathcal{S}$.
Then the environment reveals to the agent a reward $r_h$ sampled from $R_h(\cdot|s_h,a_h)$ and transitions to the next state $s_{h+1}$ following  $\prob_h^S(\cdot|s_h,a_h)$. The episode ends when the timestamp reaches $H+1$.

The agent takes actions according to some deterministic policy. A history-dependent deterministic policy $\psi=\{\psi_h \}_{h\in[H]}$ is defined as a sequence of mappings $\psi_h: G_h \rightarrow \calA$, where $G_h:=\calS^h\times\calA^{h-1}\times [0,1]^{h-1}$ is the space of the history at timestamp $h$. Namely, 
$\psi_h(g_h)$ gives guidance on how the agent will select her action under a given trajectory of the history $g_h=(s_1,a_1,r_1,\cdots,s_{h-1},a_{h-1},r_{h-1},s_h)\in G_h$ at timestamp $h$. For notational convenience, let $\Psi$ denote the set of all history-dependent deterministic policies.

We assume that the agent's preference can be represented by a continuous and strictly increasing utility function $U: [0, H] \rightarrow \R$. Without loss of generality, we assume $U(0)=0$. The goal of the agent is to find a policy that maximizes her expected utility from any starting state. That is, 
\begin{eqnarray}
\label{equ:objective}
\psi^*\in\text{argmax}_{\psi\in \Psi} \mathbb{E}^{\psi}\left.\left[U\left(\sum_{h=1}^H r_{h}\right)\right|s_1=s\right], 
\end{eqnarray}
for each $s\in \mathcal{S}$, where $\E^{\psi}$ denotes the expectation under both the transition kernel and the policy $\psi$. Note that for a general utility function $U$, the optimal policy $\psi^*$  may be history-dependent, in contrast to   Markovian policy which relies solely on the current state \citep{bauerle2014more}.  This suggests that it requires more information than the current state to make the optimal decision.

Distinct types of utility functions yield different characteristics in optimal policies. For instance, when the agent possesses a risk-neutral utility function—linear in nature—and finite state and action spaces along with a finite horizon, there exists an optimal policy that is Markovian \citep{azar2017minimax,sidford2018near}. Another special case arises when the agent's utility function is exponential. B\"auerle and Rieder demonstrated that, under mild assumptions, there always exists an optimal policy that is Markovian \citep{bauerle2014more,fei2020risk}. However, non-Markovian optimal policies often emerge for more general utility functions. 



A few popular choices of utility functions include the 
exponential utility function, the Constant Relative Risk Aversion (CRRA) utility, and the Hyperbolic absolute risk aversion (HARA) utility \citep{duffie1997hedging,fei2021exponential}. With an exponential utility $U(y)=\frac{1-e^{-\beta y}}{\beta},$
the agent is {\it risk-seeking} if $\beta<0$ and {\it risk-averse} if $\beta>0$. $\beta=0$ recovers the case where the agent is {\it risk-neutral}. With a CRRA utility function 
$U(y)=\frac{1}{1-\gamma}y^{1-\gamma}$,  the agent is {\it risk-averse} if $\gamma>0$.

\subsection{Enlarged State Space Approach}\label{sec:enlarged_state_space}
To facilitate the idea of DPP, we construct another MDP with an enlarged state space by adding a new dimension that accounts for the cumulative reward. With the enlarged state space, one can show that the optimal policy depends only on the current state $s\in \calS$ and the current cumulative reward, independent of how past rewards were collected. Hence, this enlarged state forms a {\it sufficient statistic} to make optimal decisions \citep{bauerle2014more,haskell2015convex,isohatala2023dynamic,kreps1977decision}.

\begin{definition}[Enlarged State Space and Transition Kernel]
\label{def:ERSMDP}
Given a risk-sensitive MDP $\MDP$,  the corresponding enlarged state space and transition kernel are defined as:
\begin{itemize}
\item  The collection $\calY:=\{\calY_h\}_{h\in[H]}$ is defined as a set of cumulative reward spaces with $\calY_h {:=} [0,h-1]$ being the space of possible cumulative reward at the start of timestamp $h$.  We define $\calS\times \calY_h$ as the enlarged state space at timestamp $h$, and denote $(s,y)\in \calS\times {\calY_h}$ as an enlarged state.  
\item $\prob = \{\prob_h\}_{h\in [H]}$ is defined as a set of probability transitions over the enlarged state space. Particularly, $\prob_h : \calS\times {{\calY_h}} \times \calA \rightarrow \calP(\calS\times {{\calY_h}})$ denotes the probability transition at timestamp $h$ so that 
$\prob_h(s^\prime,y^\prime|s,y,a) := \prob_h^S(s^\prime|s,a)R_h(y^\prime-y|s,a)$.
\end{itemize}
\end{definition}
Therefore, the cumulative reward state follows $y_{h+1}=\sum_{h'=1}^h r_{h'}$ with $r_{h'}\sim R_{h'}(\cdot|s_{h'},a_{h'})$. 
With the above setup for the enlarged state space, we now define the Markovian policy, the value function, and the Bellman operator.
\begin{definition}[{Markovian Policy, Value Function and Bellman Operator}]
Given a $\RSMDP$, we define its corresponding Markovian policy, value function, and Bellman operator below.
\begin{itemize}
\item A deterministic Markovian policy $\pi:=\{\pi_h\}_{h\in[H]}$ is defined by a sequence of mappings $\pi_h:\calS\times{\calY_h}\rightarrow\calA$. Let $\Pi$ denote the set of all deterministic Markovian policies for $\texttt{RS-MDP}$. Note that the set of Markovian policies is a subset of history-dependent policies.
\item The value function $V_h^\pi(\cdot,\cdot): \calS \times {\calY_h} \rightarrow \mathbb{R}$ is defined as the expected utility of the cumulative reward following the Markovian policy $\pi\in\Pi$  and starting from timestamp $h$ and state $(s,y)$. That is:
\begin{equation}
V_h^\pi(s,y):=\mathbb{E}^{\pi}\left.\left[U\left(\sum_{h^\prime=h}^H r_{h^\prime}+y\right)\,\right| \,s_h=s,y_h=y\right],
\end{equation} 
with terminal condition $V_{H+1}^\pi(s,y):=U(y)$ for all $(s,y)\in\calS\times\calY_{H+1}$. 
\item The Q function $Q_h^\pi(\cdot,\cdot,\cdot): \calS \times {\calY_h} \times \calA \rightarrow \calR$ is defined as the expected utility of the cumulative reward, by taking action $a$ at state $(s,y)$ at timestamp $h$ and following the Markovian policy $\pi\in\Pi$  thereafter. That is:  
\begin{equation}
Q_h^\pi(s,y,a):=\mathbb{E}^{\pi}\left.\left[U\left(\sum_{h^\prime=h}^H r_{h^\prime}+y\right)\,\right| \, s_h=s,y_h=y,a_h=a\right].\end{equation}


\item For each timestamp $h\in[H]$, the Bellman operator for policy $\pi\in\Pi$, denoted as $\calT^\pi_h$, is defined as:
\begin{eqnarray}
    \calT^\pi_h V(s,y) := \E^\pi\Big[V(s^\prime,y+r_h)\,\Big|\, s_h=s,y_h=y \Big].
\end{eqnarray}

\item For each timestamp $h\in[H]$, the Bellman operator  for the optimal policy $\calT_h$ is defined as:
\begin{eqnarray}
\calT_h V(s,y) := \displaystyle \mathop{\max}_{a\in \calA}\ \E\Big[V(s^\prime,y+r_h)\Big| s_h=s,y_h=y \Big].
\end{eqnarray}
\end{itemize}
\end{definition}

When the model is fully known, we have the following result showing that the Bellman equation holds and the optimal policy is Markovian. This facilitates RL algorithmic design in the next section. The result is largely inspired by Theorem 3.1 in \citet{bauerle2014more}, in which the reward functions are assumed to be deterministic. 
 {\color{black}For completeness of the paper, we provide the proof for Theorem \ref{thm:markovian_policy} in Appendix \ref{sec:append_A1} } with techniques different from \citet{bauerle2014more} but closely follow Theorem 4.4.2 in \citet{puterman2014markov} under a more elementary argument. 
\begin{theorem}[Bellman Optimality Conditions and Markovian Policy Equivalence] \label{thm:markovian_policy} Let $\pi\in\Pi$ be a Markovian policy for $\RSMDP$ {on the enlarged state space}.  Under the condition that $U$ is continuous and strictly increasing,  the following results hold:
\begin{enumerate}[label=(\alph*)]
\item For each $h\in[H]$, the Bellman equation holds that:
\begin{eqnarray}
   V_h^\pi = \calT^\pi_h V_{h+1}^\pi, 
\end{eqnarray}
with boundary condition $V_{H+1}^\pi(s,y):=U(y)$.
\item 
Let $J_1^*(s)$ denote the optimal value function following some history-dependent policy in $\Psi$ from initial state  on the original state space $s\in\calS$: 
\begin{equation}
\label{equ:optimality}
J_1^*(s):=\sup_{\psi\in\Psi}
\mathbb{E}^{\psi}\Big[U\big(\sum_{h=1}^H r_{h}\big)\,\Big|\, s_1=s\Big].
\end{equation}
Then the Bellman optimality condition holds that:
\begin{eqnarray}
[\calT_1\calT_2\cdots\calT_H U](s,0)= J_1^*(s).
\end{eqnarray}
\item Define an optimal policy $\psi^*\in\Psi$ as a policy such that $J_1^*(s) =  \mathbb{E}^{\psi^*}\Big[U\big(\sum_{h=1}^H r_{h}\big)\,\Big|\, s_1=s\Big]$. Then there exists an optimal policy $\pi^*=\{\pi^*_h\}_{h=1}^H\in \Pi$  which is Markovian. In addition, for each $h\in[H]$, if we define 
\begin{equation}
\label{equ:policy_tranlation}
\psi^*_h(g_h):=\pi^*_h\left(s,\sum_{{h^\prime}=1}^{h-1} r_{h^\prime}\right),
\end{equation}

then $\psi^*=\{\psi^*_h\}_{h\in[H]}$ is an optimal history-dependent  policy for $\MDP$ that solves \eqref{equ:objective}. 
\end{enumerate}
\end{theorem}

Theorem \ref{thm:markovian_policy} suggests that in order to find an optimal policy for $\MDP$ under the original formulation \eqref{equ:objective} which may be history-dependent, it is sufficient to find an optimal Markovian policy on the enlarged state space. {\color{black}Consequently, define the optimal Markovian policy $\pi^*:=\{\pi_h\}_{h=1}^H\in \Pi$ as:
\begin{equation}
\pi^*\in\argmax_{\pi\in\Pi}V_1^\pi(s,0), \text{ for all }s\in\calS.
\end{equation}
For the optimal Markovian policy $\pi^*:=\{\pi^*_h\}_{h=1}^H$ defined above, let us define $\{\pi^*_s\}_{s=h}^H$ as the subset of this policy class starting from time-stamp $h$. By the DPP, we have 
\begin{eqnarray}
    \{\pi^*_s\}_{s=h}^H \in\argmax_{\pi\in\Pi}V_h^\pi(s,y), \text{ for all }s\in\calS \text{ and } y\in\mathcal Y.
\end{eqnarray}
Namely, $\pi^*$ is time consistent and optimal for all time $h\in[H]$.
}

For the rest of this section, we provide a few analytical properties of $\RSMDP$, which facilitate efficient algorithm design in the next section. To start, we observe that the transition over the enlarged state space has the following structural property.
\begin{proposition}[Structural Properties of RS-MDP]
\label{par:property}
Given $\RSMDP$,  the transition kernel $\prob$ satisfies the following  property:
    $$\prob_h(s^\prime,y^\prime\mid s,y,a)=\prob_h(s^\prime,y^\prime+c\mid s,y+c,a), \quad  \forall c\in \mathbb{R} \,\,{\rm s.t.}\,\, y', y'+c\in \mathcal{Y}_h.$$ 
  Namely,  $\prob_h$ is translation invariant in $y'$ and $y$. It also implies that given a state-action pair $(s,a)$, the distribution over the next state $s^\prime\in\calS$ does not depend on the current cumulative reward $y$. 
\end{proposition}
This structural property leads to the following results on the Lipschitz continuity property of the optimal value function. 
\begin{theorem}
\label{thm:Lipschitz}
Given $\RSMDP$, if $U(\cdot)$ is $\kappa$-Lipschitz, then the optimal value function at a given state $V_h^{{\pi^*}}(s,\cdot)$, associated to the optimal Markovian policy $\pi^*$, is also $\kappa$-Lipschitz.
\end{theorem}

\begin{remark}[Remark of Theorem \ref{thm:Lipschitz}]
{
\label{remark:4}
Theorem \ref{thm:Lipschitz} suggests that, with a Lipschitz continuous utility function, the value function of the optimal Markovian policy is also Lipschitz continuous in the cumulative reward. It is, however, not the case for a generic Markovian policy. The reason is that given $y_1$ and $y_2$ close enough, the action $\pi_h(s,y_1)$ and $\pi_h(s,y_2)$ might be very different, leading to no continuity in the value function.

To illustrate this subtlety, the following example demonstrates that there exists a policy for which the value function is not Lipschitz. Suppose there is an RS-MDP with action set $\calA=\{\text{good action}, \text{bad action}\}$, where the good action yields a high reward, while the bad action yields a low reward. Suppose the state transition does not depend on the action, that is, for any $s\in\calS$, for any $h\in[H]$, 
$$P_h^S(\cdot|s,{\text{good action}})=P_h^S(\cdot|s,{\text{bad action}}).
$$  Consider the following policy $\pi=\{\pi_h\}_{h\in[H]}$: for any $h\in[H]$, for any $s\in\calS$, 
\begin{align*}
\pi_h(s,y)=\begin{cases}
\text{good action},& \text{if y is a rational number},\\
\text{bad action},& \text{if y is an irrational number}.
\end{cases}
\end{align*}
By simple calculations, one can show that the value function of this policy is not Lipschitz. Indeed, if we pick $y_1$ and $y_2$ close enough with $y_1$ being a rational number while $y_2$ being an irrational number, it holds that
\begin{align}
&|V_{H}^\pi(s,y_1)-V_H^\pi(s,y_2)|\nonumber\\ 
&=|\E_{r_1\sim R_H(\cdot|s,\text{good action})}[U(y_1+r_1)]-\E_{r_2\sim R_H(\cdot|s,\text{bad action})}[U(y_2+r_2)]|.\label{equation_theorem_Lipschitz_1}
\end{align}
As $|y_1-y_2|$ goes to zero, \eqref{equation_theorem_Lipschitz_1} does not go to zero. Therefore, $V_H^\pi(s,y)$ is not Lipschitz in $y$.}
\end{remark}

{\color{black}
With Theorem \ref{thm:Lipschitz}, we can discretize the environment, ensuring that the error scales according to the Lipschitz coefficient (see Proposition \ref{lemma:Lipschitz_Vbar_pihat}). The goal for environment discretization is to handle the continuity of the augmented state space for ``cumulative rewards''. This theorem guides our design of discretization for the algorithm implementation, with further details provided in the next subsection. The proof of Theorem \ref{thm:Lipschitz} is deferred to Appendix \ref{sec:append_A2}.

With the example in Remark \ref{remark:4}, we emphasize the subtlety of the Lipschitz property.  Theorem \ref{thm:Lipschitz}  applies exclusively to the optimal Markovian policy. Although the utility function is Lipschitz,  it is not necessary that $V_h^{{\pi}}(s,\cdot)$ is Lipschitz for an {\it  arbitrary} policy $\pi$. To overcome this, we developed a near-Lipschtizness property for near-optimal policies {(see Proposition \ref{lemma:Lipschitz_Vbar_pihat})}, which facilitates the theoretical development of our proposed RL algorithms in Section \ref{sec:VIGU} and Section \ref{sec:VIGU-UCB_simulator}.}

\subsection{Discretized Environment}
\label{sec:discretized_env}

\label{sec:FSA}
Without structural assumptions on the MDP (such as linear MDP \citep{jin2020provably,yang2019sample}), the convergence or sample complexity of existing RL algorithms heavily relies on the cardinality of the state space \citep{azar2017minimax,jin2018q}. {Since the state space $\mathcal{Y}$ is continuous,} existing methods are not directly applicable to the setting in  Section \ref{sec:enlarged_state_space}. To address this challenge, two possible methods are covering net {(e.g. \citet{wang2020reinforcement})} and function approximation {(e.g. \citet{jin2020provably})}. In this work, we adopt the covering net method and introduce a discretized environment. With a covering net over the state space $\mathcal{Y}$ for cumulative reward, an additional approximation error is introduced. To bound this error, we develop propositions showing the ``near-Lipschitz" property for the value function of a near-optimal policy by utilizing the Lipschitz results from Theorem \ref{thm:Lipschitz}.

Let us start with an introduction of the $\epsilon_o-$covering over the space of the single-period reward and cumulative reward. 
First, recall that the one-period stochastic reward is bounded between $0$ and $1$ almost surely. Then
let $\barI:=\left\{0,\epsilon_o,2\epsilon_o,\dots, \lfloor\frac{1}{\epsilon_o}\rfloor\epsilon_o\right\}$ be an $\epsilon_o$-covering on $[0,1]$ with cardinality $|\barI|=\mathcal{O}\left(\frac{1}{\epsilon_o}\right)$. 
Also, note that the cumulative reward is bounded, that is, $y_h\in [0,h-1].$  Let $\barY_h:=\left\{0,\epsilon_o,2\epsilon_o,\dots, \lfloor\frac{h-1}{\epsilon_o}\rfloor\epsilon_o\right\}$ be an $\epsilon_o$-covering on $[0,h-1]$ with cardinality $|\barY_h|=\mathcal{O}\left(\frac{H}{\epsilon_o}\right)$. \label{definition_rtilde}
Let ${\rm Proj}_{{\barY_h}}(y)$ denote the $\ell_2$ projection of $y\in {\calY_h}$ to the set ${\barY_h}$. {\color{black}Note that since $y\in\mathbb{R}$, the $\ell_2$ and $\ell_1$ projections are equivalent.} For points that are equally distanced from two points in ${\barY_h}$, we choose the smaller one for simplicity. Similarly, let ${\rm Proj}_{\barI}(r)$ denote the $\ell_2$ projection of $r\in [0,1]$ to the set $\barI$. 

Additionally, with the introduced covering net, let us define the corresponding policy class. Namely, let $\barPi$ be the set of Markovian policies defined on the covering nets {$\{\barY_h\}_{h=1}^{H+1}$} with 
\begin{eqnarray}\label{eq:barPi}
    \barPi:=\left\{\barpi=\{\barpi_h\}_{h\in[H]}\,\,\Big|\,\, \barpi_h:\calS\times\barY_h\rightarrow\calA\right\}.
\end{eqnarray}
We define $\barR:=\{\barR_h\}_{h\in[H]}$ and $\bar\prob:=\{\barprob_h\}_{h\in[H]}$ as the ``discretized'' approximation for reward distribution $R$ and transition kernel $R$ and $\prob$ as follows. For any $h\in[H]$, define $\barR_h:\calS\times\mathcal{A}\rightarrow{\mathcal{P}(\barI)}$ and $\barprob_h:\calS\times\barY\times\calA\rightarrow\calP(\calS\times\barY)$ with
\begin{eqnarray*}
    &&\barR_h(r\mid s,a):=\int_{r-\frac{\epsilon_o}{2}}^{r+\frac{\epsilon_o}{2}}R_h({\rm d}r'\mid s,a), \,\,{\forall r \in \barI},\\
    &&\barprob_h(s^\prime,y^\prime \mid s,y,a):=\prob^S_h(s^\prime\mid s,a)\barR_h(y^\prime-y\mid s,a).
\end{eqnarray*}

Notationally, we use $\BARMDP$ to denote the {\it Discretized Environment}, and let $\MDP$ denote the {\it Original Environment}. Mathematically, the Discretized Environment can be viewed as a special RS-MDP using $\barprob$ as its transition kernel, and $\calS\times\barY$ as its state space. Under the Discretized Environment, given a policy $\barpi\in\barPi$, let $\barV^{\barpi}$ and $\barQ^{\barpi}$ respectively be the corresponding value function and Q function. 
Let ${\barpi^*}$ be the optimal policy and denote  $\barV^{\barpi^*}$ as the associated optimal value function.

\paragraph{Connection between the Discretized Environment and the Original Environment.} To connect the Discretized Environment and the Original Environment,  we define two notions of interpolations -- policy interpolation and value function interpolation. First,  we define a policy interpolation function $\Gamma:\barPi\rightarrow\Pi$: For any policy $\barpi\in\barPi$, $\Gamma(\barpi):=\{\Gamma_h(\barpi)\}_{h\in[H]}$ with 
\begin{equation}
\label{equation_Gamma}
    \Gamma_h(\barpi)(s,y):=\barpi_h(s,\projY{y}).
\end{equation}
This mapping $\Gamma$ converts any policy $\barpi\in\barPi$ into a policy in $\Pi$ by interpolation.

{\color{black}Next, we define the interpolation for value functions. Note that $\barV_h$ is a mapping defined on $\calS\times\barY_h$ and $\barQ_h$ is a mapping defined on $\calS\times\barY_h\times\calA$. For any $h\in[H+1]$,  $(s,a)\in\calS\times\calA$, and $y\in\calY_h$ but with $y\notin\barY_h$, we define the value function interpolation as:

\begin{align}
\label{equation_value_function_interpolation}
&\barV_h^\barpi(s,y):=\barV_h^\barpi(s,\projY{y}), \\
&\barQ_h^\barpi(s,y,a):=\barQ_h^\barpi(s,\projY{y},a).  
\end{align}
The value function interpolation extends the value functions $\barV_h$ and $\barQ_h$ to the (original) enlarged space $\calS\times{\calY_h}$. For  the rest of the paper, we define $\barV_h$ 
as a mapping on $\calS\times\calY_h$, and $\barQ_h$ as a mapping on $\calS\times\calY_h\times\calA$ by interpolation.}

{\color{black} Now, with the above definitions in place, we establish connections between the Discretized Environment and the Original Environment in the following propositions. Given that $\BARMDP$ is a discretized approximation of the original RS-MDP,   in the next proposition, we demonstrate that the difference between the optimal value functions of $\RSMDP$ and $\BARMDP$ can be bounded by leveraging the fact that the optimal value functions are Lipschitz. All the proofs of the propositions in this section are deferred to Appendix \ref{sec:append_B}.}

\begin{proposition}
\label{lemma:v_difference_ERSMDP_BARMDP}
The difference between the optimal value functions of $\RSMDP$ and $\BARMDP$ can be bounded by the covering distance $\epsilon_o$. Formally, for any $s\in \calS$, $y\in\barY$, $h\in[H]$, 
\begin{equation}
\label{equ:lemma_v_difference_ERSMDP_BARMDP_1}
    \left|V_h^{\pi^*}(s,y)-\barV^{{\barpi^*}}_h(s,y)\right|\leq (H-h+1) \kappa \epsilon_o.
\end{equation}
\end{proposition} 

From Theorem \ref{thm:Lipschitz} we know that the optimal policy is $\kappa$-Lipschitz. Indeed, we can show that a near-optimal policy also has a certain ``near-Lipschitz'' property in the following proposition.

\begin{proposition}[``Near-Lipschitz" Property of a Near-optimal Policy]
\label{lemma:Lipschitz_Vbar_pihat}
Take some Markovian policy $\barpi\in\barPi$ that is near-optimal under the Discretized Environment in the sense that, for any $h\in[H]$, {\color{black} there exists  $\Delta_h\in\mathbb{R}$ such that} 
\begin{align}\label{equ:lemma_Lipschitz_Vbar_pihat_0}
\barV_h^{\barpi^*}(s,y)-\barV_h^{\barpi}(s,y)\leq \Delta_h, \quad \forall (s,y)\in \calS\times\barY_h.
\end{align}
Then we have for any $h\in[H]$, {\color{black} it holds that} 
\begin{equation}
\label{equ:lemma_Lipschitz_Vbar_pihat_1}
\Big|\barV^\barpi_h(s,y_1)-\barV^\barpi_h(s,y_2)\Big|\leq 2\Delta_h+\kappa\left|\projY{y_1}-\projY{y_2}\right|,
\end{equation}
for any $s\in\calS$, and $y_1,y_2\in{\calY_h}$. 
\end{proposition}

Proposition \ref{lemma:Lipschitz_Vbar_pihat} indicates that a near-optimal policy has some ``near-Lipschitz'' property under $\BARMDP$. 
Since $\BARMDP$ is a discretized approximation of $\RSMDP$,  we are able to bound the difference between the value functions of a near-optimal policy under $\RSMDP$ and $\BARMDP$, by utilizing this ``near-Lipschitz'' property. 
\begin{proposition}
\label{lemma:V_Vbar_difference}
Take some Markovian policy $\barpi\in\barPi$ that is near-optimal under the Discretized Environment in the sense that, for any $h\in[H]$, {\color{black} there exists  $\Delta_h\in\mathbb{R}$ such that}   
\begin{align}
\barV_h^{\barpi^*}(s,y)-\barV_h^{\barpi}(s,y)\leq \Delta_h, \quad (s,y)\in \calS\times\barY_h.
\end{align}
{\color{black} Then we have }for any $h\in[H+1]$ and $(s,y)\in\calS\times{\calY_h}$, {\color{black}it holds that}
\begin{equation}
\Big|V_h^{\Gamma(\barpi)}(s,y)-\barV_h^\barpi(s,y)\Big| \leq 2\sum_{h'=h}^H \Delta_{h'} + (H-h+1)\kappa\epsilon_o.
\label{equ:lemma_V_Vbar_difference_1}
\end{equation}
\end{proposition}

{Proposition \ref{lemma:V_Vbar_difference} shows that as long as we find a near-optimal policy under the Discretized Environment, the converted policy will also be a near-optimal policy under the Original Environment. Therefore, in order to develop an efficient algorithm, it suffices to find a near-optimal policy under the Discretized Environment.}

\section{Algorithm Design and Finite Sample Analysis with Simulator}\label{sec:VIGU}
 In the context of RL, the agent does not know the transition kernels $\prob^S$ nor the distribution of the reward $R$. Instead, the agent aims to improve her choice of actions (to optimize certain objectives) by interacting with the unknown environment.  At the beginning of each timestamp $h$, the
agent observes the state of the environment $s_h \in \mathcal{S}$ and selects an action $a_h \in \mathcal{A}$. At
the end of this timestamp, in part as a consequence of its action, the agent receives a stochastic reward
$r_h$  and a new state $s_{h+1}$ from the environment. The tuple $(s_h, a_h, r_h, s_{h+1})$ is called
a sample at timestamp $h$. 

With access to a simulator $\calG$ for  $\MDP$,  the agent is able to collect the next state and reward from {\it any} state, action, and timestamp. Namely, $(s^\prime,r)=\calG(s,a,h)$ returns a sample with $s^\prime \sim \prob_h^S(\cdot\mid s,a)$ and $r \sim R_h(\cdot\mid s,a)$. 

The agent's objective is to find a near-optimal policy while minimizing the number of samples acquired from the environment (or simulator). To provide a more mathematically precise evaluation criterion, we introduce the concepts of an $\epsilon$-optimal {\color{black}Markovian} policy and sample complexity as follows.
\begin{definition}[$\epsilon$-optimal Markovian Policy]
A Markovian policy $\pi$ is called an $\epsilon$-optimal policy if for any $h\in [H]$,
\begin{equation}\label{eq:def_epsilon_Markovian}
    \infnorm{V_h^{\pi^*}-V_h^{\pi}} \leq \epsilon.
\end{equation}
\end{definition}
{\color{black} Note that it is sufficient to work with the $\epsilon$-optimal Markovian policy as defined in \eqref{eq:def_epsilon_Markovian}. This is because one can show that an $\epsilon$-optimal and history-dependent policy for $\texttt{RS-MDP}$ can be converted into an $\epsilon$-optimal  Markovian policy using \eqref{equ:policy_tranlation}. } 
 
\begin{definition}[Sample Complexity]
Given $\epsilon>0$ and $0<p<1$, the sample complexity for finding an $\epsilon$-optimal policy of a given algorithm is the minimum number 
of samples $M$ such that with probability at least $1-p$, the output policy using $M$ samples is $\epsilon$-optimal.
\end{definition}

\subsection{Value Iteration Method with Access to the Simulator}

We propose a value iteration method with $\epsilon_o-$covering, which is detailed in Algorithm \ref{alg1}. The algorithm contains two steps. In the first step (lines 2-6), we estimate the transition kernel $\mathbb{P}_h^S$ and the reward distribution $R_h$. In the second step (lines 7-16), we estimate the Q function based on the estimated kernels and then compute the corresponding optimal policy.
The analysis of the algorithm is based on the properties of the Discretized Environment introduced in Section \ref{sec:discretized_env}. {\color{black}Note that the covering net is only used on the augmented state space of the cumulative rewards and therefore the proposed algorithm is computationally efficient {\color{black}(with a time complexity that is polynomial in $S$, $A$, $H$, $T$, and $\kappa$)}}.

\begin{algorithm}[h]
\caption{Value Iteration under General Utility (VIGU) 
\label{alg1}}
\begin{algorithmic}[1]
\State Initialize: $\forall s,y \in \calS \times \barY_{H+1},\ \hat{V}_{H+1}(s,y):= U(y)$.
\For{$(s,a,h) \in \calS \times \calA \times [H]$}
\State Query the simulator $\calG(s,a,h)$ for $n$ times and collect {\bf iid} samples $\{(s^1,r^1),(s^2,r^2)\cdots (s^n,r^n)\}$. 
\State \label{line:definition_ps_hat} Estimate the transition $\hat{\prob}^S_h(s^\prime \mid s,a)$ for all $s^\prime \in \calS$: 
\begin{eqnarray*}
    \hat{\prob}^S_h(s^\prime \mid s,a)=\frac{\sum_{i=1}^n \mathbbm{1}\{s^i=s^\prime\}}{n}.
\end{eqnarray*}
\State Estimate the transition $\hat{R}_h(r \mid s,a)$ for all $r \in \barI$: 
\label{line:definition_r_hat}
\begin{eqnarray*}
    \hat{R}_h(r \mid s,a)=\frac{\sum_{i=1}^n \mathbbm{1}\{{\rm Proj}_{\barI}(r^i)=r\}}{n}.
\end{eqnarray*}
\EndFor
\For{$h = H,H-1,\cdots, 1$}
\For{$(s,y) \in \calS \times \barY$}
\For{$a \in \calA$}
\State Compute the estimation of the Q function:
\label{line:VIGU_Qhat_update}
\begin{eqnarray*}
    \hat{Q}_h(s,y,a) = \E_{s^\prime \sim \hat{\prob}^S_h(\cdot\mid s,a),
    r \sim \hat{R}_h(\cdot\mid s,a)}\left[\hat{V}_{h+1}(s^\prime,y+r)\right].
\end{eqnarray*}
\label{line:update}
\EndFor
\State Define $\hat{V}_h(s,y) = \max_a  \hat{Q}_h(s,y,a)$. 
\State Set $\hatpi_h(s,y) \in arg\max_a \hat{Q}_h(s,y,a)$. 
\EndFor
\EndFor
\State \textbf{Output:} ${\Gamma(\hatpi)}$. $\Gamma$ is the policy interpolation function defined in \eqref{equation_Gamma}.
\end{algorithmic}
\end{algorithm}
In Algorithm \ref{alg1},  estimating $\prob^S_h(\cdot\mid s,a)$ and $R_h(\cdot\mid s,a)$ separately is equivalent to estimating $\prob_h(\cdot,\cdot\mid s,y,a)$ thanks to Proposition \ref{par:property}. More specifically, Line \ref{line:update} is equivalent to computing $\hat{Q}_h(s,y,a) = \E_{(s^\prime,y^\prime)\sim \hat{\prob}_h(\cdot,\cdot\mid s,y,a)}\left[\hat{V}_{h+1}(s^\prime,y^\prime)\right]$, where the (joint) empirical  distribution  $\hat{\prob}_h(s^\prime,y^\prime \mid s,y,a )$ is defined as: 
\begin{eqnarray}
\label{equ:definition_p_hat}
\hat{\prob}_h(s,y^\prime \mid s,y,a ):=\hat{\prob}^S_h(s^\prime \mid s,a)\hat{R}_h(y^\prime-y \mid s,a).
\end{eqnarray}
We then define $\hatprob:=\{\hatprob_h\}_{h\in[H]}$ as the estimated transition kernels from Algorithm \ref{alg1}.
\subsection{Theoretical Analysis}
In this section, we provide the sample complexity analysis of Algorithm \ref{alg1}.
\begin{theorem}
\label{thm:bound}
Let $\pi^*\in\Pi$ be the optimal Markovian policy for $\RSMDP$ and let ${\Gamma(\hatpi)}=\{{\Gamma(\hatpi)}_h\}_{h=1}^H$  be the output of Algorithm \ref{alg1}. Under the condition that $U(\cdot)$ is $\kappa$-Lipschitz, we have with probability at least $1-p$, for any $h\in [H]$, 
\begin{eqnarray}
    \infnorm{V_h^{\pi^*}-V^{{\Gamma(\hatpi)}}_h} 
    := \sup_{s\in \calS, y\in {\calY_h}} \left | V_h^{\pi^*}(s,y)-V^{{\Gamma(\hatpi)}}_h(s,y) \right |= \calO \left(H\kappa   \epsilon_o +H^3\kappa \sqrt{\frac{{\iota_1}}{n}}\right),
\end{eqnarray}
where ${\iota_1}=\log \Big(\frac{4H^2SA}{p\epsilon_o}\Big)$ and
$\infnorm{\cdot}$ denotes the infinity norm. Note that $p$ is an input since it enters through $\iota_1$.
\end{theorem}
Theorem \ref{thm:bound} bounds the error of the value functions between the output policy from Algorithm \ref{alg1} and the optimal policy. The result  can be converted to the number of samples that are required to find a near-optimal policy. As a consequence of Theorem \ref{thm:bound},  we have the following corollary bounding the sample complexity.
\begin{corollary}
\label{corollary:sample}
Let $\epsilon_o = \Theta\left(\frac{\epsilon}{H\kappa }\right)$ and $n=\Theta\left(\frac{ H^6\kappa^2{\iota_1}}{\epsilon^2}\right)$, then
the corresponding sample complexity for finding an $\epsilon$-optimal policy with probability at least $1-p$ is 
\begin{eqnarray}
\mathcal{O}\left(\frac{H^7SA\kappa^2 {\iota_1}}{\epsilon^2}\right),
\end{eqnarray}
where ${\iota_1}$ is defined in Theorem \ref{thm:bound}.
\end{corollary}
The lower bound for the risk-neutral utility function, which is a special case of our setting, is $\Omega\left(\frac{H^4SA}{\epsilon^2}\right)$ \citep{azar2017minimax,agarwal2020model}. Our result matches this lower bound up to a cubic factor in $H$, a quadratic factor in the Lipschitz constant $\kappa$, and a logarithm factor {of all model parameters}. 

One reason for the difference in $H$ between the above lower bound and our upper bound in the sample complexity is that we assume the agent receives {\it continuous} stochastic rewards instead of deterministic rewards. In the risk-neutral setting, the key difficulty lies in efficiently estimating the transition kernel $\prob^S_h: \mathcal{S}\times \mathcal{A} \rightarrow \calP(\mathcal{S})$ \citep{jin2018q,yang2019sample}. However, in our risk-sensitive setting, compared to estimating $\prob_h^S$, 
it is more challenging to estimate the distribution $R_h:\calS\times\calA\rightarrow\calP([0,1])$. This is because the stochastic reward considered here is a continuous random variable instead of a discrete one. 

\subsection{Proof of Theorem \ref{thm:bound}}
\label{subsection_proof_theorem_3}
To start, we first introduce some additional notations.  

\paragraph{Output Environment from the Algorithm.} Recall that $\hatprob$ is the estimated distribution calculated from Algorithm \ref{alg1} (see Line \ref{equ:definition_p_hat}). Let $\HATMDP$ denote the {\it Output Environment}, that is, the RS-MDP under $\hatprob$. Note that $\hatV$ and $\hatQ$ calculated in Algorithm \ref{alg1} (see Line \ref{line:VIGU_Qhat_update}) satisfy the optimal Bellman equations under the Output Environment and therefore respectively are the associated optimal value function and Q function. The output policy ${\Gamma(\hatpi)}$ is the greedy policy according to $\hatQ$ and therefore is the optimal policy under the Output Environment. 

Given a sample $r\sim R_h(\cdot\mid s,a)$, for any $r_o\in\barI$ we have
\begin{align} 
R_h\Big(\projR{r}=r_o\mid s,a \Big)=&R_h\left(r_o-\frac{\epsilon_o}{2} < r\leq r_o+\frac{\epsilon_o}{2} \,\Big|\, s,a \right)\nonumber\\
    =&\int_{r_o-\frac{\epsilon_o}{2}}^{r_o+\frac{\epsilon_o}{2}} R_h({\rm d}r'\mid s,a) \nonumber\\
    =&\barR_h(r_o\mid s,a).\label{section_VIGU_analysis_equation_1}
\end{align}
Therefore, given $r\sim R_h(\cdot\mid s,a)$, we have $\projR{r}\sim\barR_h(\cdot\mid s,a)$. In Algorithm \ref{alg1}, $\hatR$ is constructed using projected samples $\projR{r^i}$ (see Line \ref{line:definition_r_hat}). Therefore, $\hatR$ is an unbiased estimator of $\barR$. $\hatprob^S$ in Line \ref{line:definition_ps_hat} is {an} unbiased estimator of $\prob^S$, and hence $\hatprob$ in \eqref{equ:definition_p_hat} is {an} unbiased estimator of $\barprob$. By using concentration inequalities, we will show that ${\Gamma(\hatpi)}$ is a near-optimal policy under $\BARMDP$. For any $(s,y)\in\calS\times\barY$, the difference  between $\barV_h^{\Gamma(\hatpi)}(s,y)$ and $\hatV_h^{\Gamma(\hatpi)}(s,y)$ will be bounded formally in Lemma \ref{lemma:performance_difference}. \\

In summary, we have three RS-MDPs and their corresponding optimal value functions:
$V_h^{\pi^*}$ is the optimal value function for $\RSMDP$, $\barV_h^{{\barpi^*}}$ is the optimal value function for $\BARMDP$, $\hatV_h^\hatpi$ is the optimal value function for $\HATMDP$. The connection of these value functions will be further explained in the following lemmas. 

Next, we show that for any policy $\barpi\in\barPi$, the difference between the associated value functions under $\BARMDP$ and $\HATMDP$ can be bounded using concentration inequalities. To ease the notation, let us define
\begin{eqnarray*}
    [\prob_h V_{h+1}](s,y,a):=\E_{(s',y')\sim\prob_h(\cdot\mid s,y,a)}[V_{h+1}(s',y')].
\end{eqnarray*}
\begin{lemma}[Performance Difference]
\label{lemma:performance_difference}
Given any policy $\barpi\in\barPi$, 
for any $(s,y)\in\calS\times\barY$,
\begin{equation}\label{equation:lemma_performance_difference_1}
    \barV_h^{\barpi}(s,y)-\hat{V}_h^{\barpi}(s,y) = \left.\E^{\barpi}_{\hat{\prob}} \left[\sum_{h'=h}^H(\barprob_{h'}-\hat{\prob}_{h'} )\barV_{h'+1}^{\barpi}(s_{h'},y_{h'},a_{h'}) \right| s_h=s,y_h=y  \right ],
\end{equation}
and with probability at least $1-\frac{p}{2}$, $\forall h'\in[H]$, $\forall (s,y)\in\calS\times\barY$,
\begin{equation}
\left|\barV_h^{\barpi}(s,y)-\hat{V}_h^{\barpi}(s,y) \right|\leq \kappa(H-h+1)H\sqrt{\frac{{\iota_1}}{n}},
\end{equation}
where ${\iota_1}$ is defined in Theorem \ref{thm:bound}. Here we let $\E^{\barpi}_{\hat{\prob}}$ denote the expectation under $\HATMDP$ following policy $\barpi$. 
\end{lemma}
\begin{proof}[Proof of Lemma \ref{lemma:performance_difference}.]
For any $(s,y)\in\calS\times\barY$, 
\begin{align}
\barV_h^{\barpi}(s,y) -\hat{V}_h^{\barpi}(s,y)
&=\barQ_h^{\barpi}(s,y,\barpi_h(s,y))-\hat{Q}_h^{\barpi}(s,y,\barpi_h(s,y)) \label{equation:lemma_performance_difference_2}\\
&=\barprob_h  \barV_{h+1}^{\barpi}(s,y,\barpi_h(s,y))-\hat{\prob} _h\hat{V}_{h+1}^{\barpi}(s,y,\barpi_h(s,y)) \label{equation:lemma_performance_difference_3}\\
&= \barprob_h \barV_{h+1}^{\barpi}(s,y,\barpi_h(s,y))
-\hat{\prob}_h \barV_{h+1}^{\barpi}(s,y,\barpi_h(s,y)) \nonumber\\
&\qquad \qquad +\hat{\prob} \barV_{h+1}^{\barpi}(s,y,\barpi_h(s,y))
-\hat{\prob}_h \hat{V}_{h+1}^{\barpi}(s,y,\barpi_h(s,y))\nonumber\\
&=(\barprob_h-\hat{\prob}_h) \barV_{h+1}^{\barpi}(s,y,\barpi_h(s,y))
+\hat{\prob}_h (\barV_{h+1}^{\barpi}-\hat{V}_{h+1}^{\barpi})(s,y,\barpi_h(s,y)).\label{equation:lemma_performance_difference_4}
\end{align}
\eqref{equation:lemma_performance_difference_2} holds by the definition of the value function, and \eqref{equation:lemma_performance_difference_3} holds by applying the Bellman equation.  In \eqref{equation:lemma_performance_difference_4}, observe that 
\begin{align*}
&\hat{\prob}_h (\barV_{h+1}^{\barpi}-\hat{V}_{h+1}^{\barpi})(s,y,\barpi_h(s,y))=\E_{(s_{h+1},y_{h+1})\sim\hatprob_h(\cdot\mid s,y,\barpi_h(s,y))}[(\barV_{h+1}^{\barpi}-\hat{V}_{h+1}^{\barpi})(s_{h+1},y_{h+1})].
\end{align*}
Apply this to \eqref{equation:lemma_performance_difference_4} and we get: 
\begin{align}
(\barV_h^{\barpi} -\hat{V}_h^{\barpi})(s,y)&=
(\barprob_h-\hat{\prob}_h) \barV_{h+1}^{\barpi}(s,y,\barpi_h(s,y))
\nonumber\\ &+\E_{(s_{h+1},y_{h+1})\sim\hatprob_h(\cdot,\cdot\mid s,y,\barpi_h(s,y))}[(\barV_{h+1}^{\barpi}-\hat{V}_{h+1}^{\barpi})(s_{h+1},y_{h+1})].\label{equation:lemma_performance_difference_4.5}
\end{align}
Note that \eqref{equation:lemma_performance_difference_4.5} gives a recursive form. Namely, the relationship  in \eqref{equation:lemma_performance_difference_4.5} also holds for timestamp $h+1$: 
\begin{align}
(\barV_{h+1}^{\barpi}-\hat{V}_{h+1}^{\barpi})(s_{h+1},y_{h+1})&=(\barprob_{h+1}-\hat{\prob}_{h+1}) \barV_{h+2}^{\barpi}(s,y,\barpi_h(s,y))\nonumber\\
&+\E_{(s_{h+2},y_{h+2})\sim\hatprob_{h+1}(\cdot,\cdot\mid s,y,\barpi_h(s,y))}[(\barV_{h+2}^{\barpi}-\hat{V}_{h+2}^{\barpi})(s_{h+2},y_{h+2})].\label{equation:lemma_performance_difference_5}
\end{align}
Plug \eqref{equation:lemma_performance_difference_5} into \eqref{equation:lemma_performance_difference_4.5} and repeat this procedure, we prove that 
\begin{equation}
\label{equation:lemma_performance_difference_6}
\barV_h^{\barpi}(s,y)-\hat{V}_h^{\barpi}(s,y) = \left.\E^{\barpi}_{\hat{\prob}} \left[\sum_{h'=h}^H(\barprob_{h'}-\hat{\prob}_{h'} )\barV_{h'+1}^{\barpi}(s_{h'},y_{h'},a_{h'}) \right| s_h=s,y_h=y  \right ].
\end{equation}
Next, as explained below \eqref{section_VIGU_analysis_equation_1}, $\hatprob_h$ is an unbiased estimator of $\barprob_h$. Therefore, by applying Hoeffding's inequality, given any $h'\in[H]$,  with probability at least $1-\frac{p\epsilon_o}{2H^2SA}$,
\begin{align}
\left|(\barprob_{h'}-\hat{\prob}_{h'} )\barV_{h'+1}^{\barpi}(s_{h'},y_{h'},a_{h'})\right|\leq H\kappa \sqrt{\frac{{\iota_1}}{n}}. \label{equation:lemma_performance_difference_7}
\end{align}
Apply a union bound and we have with probability at least $1-\frac{p}{2}$,   \eqref{equation:lemma_performance_difference_7} holds for any $ h'\in[H]$ and for any $(s,y,a)\in\calS\times\barY\times\calA$. Applying \eqref{equation:lemma_performance_difference_7} to bound \eqref{equation:lemma_performance_difference_6},  we get the desired upper bound in high probability.
\end{proof}

With Lemma \ref{lemma:performance_difference}, we are ready to show that the greedy policy $\hatpi$ from Algorithm \ref{alg1} is a near-optimal policy under $\BARMDP$.
\begin{lemma}
\label{lemma:pibar_pihat_difference}
With probability at least $1-p$, for any $h\in[H+1]$ and $(s,y)\in\calS\times\barY_h$, the output policy ${\Gamma(\hatpi)}$ from Algorithm \ref{alg1} satisfies:
\begin{equation}
\label{equ:lemma_pibar_pihat_difference_1}
    0\leq\barV_h^{{\barpi^*}}(s,y)-\bar{V}_h^{\hatpi}(s,y) \leq 2H(H-h+1) \sqrt{\frac{\kappa^2{\iota_1}}{n}}.
\end{equation} 
\end{lemma}

\begin{proof}[Proof of Lemma \ref{lemma:pibar_pihat_difference}]
The first inequality in \eqref{equ:lemma_pibar_pihat_difference_1} holds due to the fact that ${\barpi^*}$ is the optimal policy under $\BARMDP$. To prove the second inequality in \eqref{equ:lemma_pibar_pihat_difference_1}, observe that by telescoping sums,
\begin{align}
&\barV_h^{{\barpi^*}}(s,y)-\bar{V}_h^{\hatpi}(s,y)=\barV_h^{{\barpi^*}}(s,y)-\hatV_h^{\barpi^*}(s,y)
+\hatV_h^{\barpi^*}(s,y)-\hatV_h^\hatpi(s,y)
+\hatV_h^\hatpi(s,y)-\bar{V}_h^{\hatpi}(s,y).\label{lemma_pibar_pihat_difference_equation_1}
\end{align}
By Lemma \ref{lemma:performance_difference}, with probability at least $1-p$, $\forall h \in[H]$, $\forall (s,y)\in\calS\times\barY$,
\begin{align}
&\barV_h^{{\barpi^*}}(s,y)-\hatV_h^{\barpi^*}(s,y)\leq (H-h+1)H\sqrt{\frac{\kappa^2{\iota_1}}{n}},\label{lemma_pibar_pihat_difference_equation_2}\\
&\hatV_h^\hatpi(s,y)-\bar{V}_h^{\hatpi}(s,y) \leq (H-h+1)H\sqrt{\frac{\kappa^2{\iota_1}}{n}}.\label{lemma_pibar_pihat_difference_equation_3}
\end{align}
As explained in the paragraph above \eqref{section_VIGU_analysis_equation_1}, $\hat{V}_h^{\hatpi}$ is the optimal value function for $\HATMDP$, and therefore 
\begin{equation}
\label{lemma_pibar_pihat_difference_equation_4}  \hatV_h^{\barpi^*}(s,y)-\hatV_h^\hatpi(s,y)\leq 0.
\end{equation}
Applying \eqref{lemma_pibar_pihat_difference_equation_2}, \eqref{lemma_pibar_pihat_difference_equation_3}, and \eqref{lemma_pibar_pihat_difference_equation_4} to bound \eqref{lemma_pibar_pihat_difference_equation_1}, we finish the proof.
\end{proof}

Finally, we are ready to prove the main result in Theorem \ref{thm:bound}.

\begin{proof}[Proof of Theorem \ref{thm:bound}.]
For any $(s,y)\in\calS\times{\calY_h}$, 
\begin{align*}
&V_h^{\pi^*}(s,y)-V_h^{{\Gamma(\hatpi)}}(s,y)\\
&= \underbrace{V_h^{\pi^*}(s,y)-V_h^{\pi^*}(s,\bary)}_{A_1} +\underbrace{V_h^{\pi^*}(s,\bary)-\barV_h^{{\barpi^*}}(s,\bary) }_{A_2}\\
&+\underbrace{\barV_h^{{\barpi^*}}(s,\bary)  -\barV_h^{\hatpi}(s,\bary)}_{A_3}+\underbrace{\barV_h^{\hatpi}(s,\bary) - \barV_h^{\hatpi}(s,y) }_{A_4}\\
&+\underbrace{\barV_h^{\hatpi}(s,y) -V_h^{{\Gamma(\hatpi)}}(s,y)}_{A_5}.
\end{align*}
Since $V_h^{\pi^*}$ is the optimal value function for $\RSMDP$, $A_1$ can be bounded by $\kappa \epsilon_o$ using Theorem \ref{thm:Lipschitz}. $A_2\leq H\kappa \epsilon_o$ by Proposition \ref{lemma:v_difference_ERSMDP_BARMDP}. 
 ${\Gamma(\hatpi)}$ is a near-optimal policy according to Lemma \ref{lemma:pibar_pihat_difference} and $A_3$ is therefore bounded by Proposition \ref{lemma:pibar_pihat_difference}.
$A_4=0$ by value function interpolation defined in \eqref{equation_value_function_interpolation}. 
$A_5$ is bounded by Proposition  \ref{lemma:V_Vbar_difference}.
Combining the above bounds gives with probability at least $1-p$, for any $(s,y)\in\calS\times{\calY_h}$,
\begin{align*}
&V_h^{\pi^*}(s,y)-V_h^{{\Gamma(\hatpi)}}(s,y)\\
&\leq \kappa\epsilon_o+H\kappa \epsilon_o+2H^2\sqrt{\frac{\kappa ^2{\iota_1}}{n}}
+2 H^3 \sqrt{\frac{\kappa^2{\iota_1}}{n}}+(H+1)\kappa \epsilon_o\\
&=\calO\left(
H^3 \sqrt{\frac{\kappa^2{\iota_1}}{n}}+H\kappa \epsilon_o
\right).
\end{align*}
This finishes the proof.
\end{proof}
\begin{proof}[Proof of Corollary \ref{corollary:sample}]
Let $\epsilon_o = \Theta\left(\frac{\epsilon}{H\kappa }\right)$ and $n=\Theta\left(\frac{ H^6\kappa^2{\iota_1}}{\epsilon^2}\right)$, then we have with probability at least $1-p$, for any $(s,y)\in\calS\times{\calY_h}$,
\begin{align*}
&V_h^{\pi^*}(s,y)-V_h^{{\Gamma(\hatpi)}}(s,y)\\
&=\calO\left(
H^3 \sqrt{\frac{\kappa^2 {\iota_1}}{n}}+H\kappa \epsilon_o
\right)\\
&=\calO\left(\epsilon\right).
\end{align*}

The corresponding sample complexity for finding an $\epsilon$-optimal policy, with probability at least $1-p$, is 
\begin{eqnarray}
HSAn=\mathcal{O}\left(\frac{ H^7SA\kappa^2{\iota_1}}{\epsilon^2}\right).
\end{eqnarray}
\end{proof}
\section{Algorithm Design and Regret Analysis without Simulator}\label{sec:VIGU-UCB_simulator}
In this section, we consider a more practical setting where the agent does not have access to a simulator but instead collects samples in an episodic setting. In each episode $k$, the environment provides an initial state $s_1^k$ according to some distribution, and the agent collects samples $(s_h^k,a_h^k,r_{h}^k,s_{h+1}^k)$ according to some policy, for the entire trajectory from timestamp $0$ to timestamp $H$ without restart. 

In an episodic setting, the agent tries to maximize the expected utility function of the entire episode by interacting with the environment. Therefore, it is more natural to evaluate the performance using the notation of regret, which is the difference of expected rewards collected between the optimal policy and a sequence of policies $\{\tilde{\pi}^k\}_{k=1}^K$ specified by the agent. The total (expected) regret is defined as:\begin{equation}
\label{definition_regret}
{\rm Regret}(K):=\sum_{k=1}^K V_1^{\pi^*}(s_1^k,0)-V_1^{\tilde{\pi}^k}(s_1^k,0).
\end{equation}
Note that the regret can be converted to sample complexity in the sense of the average policy of past trajectories. Details can be found in Section 3.1 in \citet{jin2018q}.

In the following sections, we apply the covering net method introduced in Section \ref{sec:FSA}. We first demonstrate that the regret under the Discretized Environment can be converted to the regret under the Original Environment under an additional assumption on the reward distribution, and then develop a value iteration algorithm using samples along the trajectory within each episode. In addition, we show that our algorithm achieves polynomial regret in the episode length $H$ and number of episodes $K$.

\subsection{Preliminary Results on {Discretized} Episodic MDP}
In this section, we illustrate some algorithm-agnostic theoretical results connecting the Discretized Environment and the Original Environment. These results will inspire the design of an efficient algorithm which will be introduced in Section \ref{sec:VIGU-UCB}.
To begin with, we propose an additional Lipschitz assumption on the reward distribution, which is a standard assumption in the stochastic control literature \citep{pham2009continuous}.
\begin{assumption}[Regularity Assumption on the Reward Distribution]
\label{assumption_lipschitz_continuous_bounded}
We assume that for each $(s,a)\in \mathcal{S}\times \mathcal{A}$ and $h
\in[H]$, the reward distribution $R_h(\cdot|s,a)\in \mathcal{P}([0,1])$ has a probability density function $f_h^R(\cdot|s,a)$ such that $R_h(\mathcal{B}|s,a) = \int_{\mathcal{B}}f_h^R(r|s,a){\rm d} r$ for any Borel set $\mathcal{B}\subseteq[0,1]$. In addition, assume that for each $(s,a)\in \mathcal{S}\times \mathcal{A}$ and $h\in[H]$, $f_h^R(\cdot|s,a)$ is $\lambda$-Lipschitz and upper-bounded by some constant $\eta\in\mathbb{R}$. That is, for any $h\in[H]$ and for any $(s,a)\in\calS\times\calA$,
\begin{eqnarray*}
|f_h^R(r\mid s,a) - f^R_h(r'\mid s,a)|&\leq& \lambda |r-r'|, \quad \forall r,r'\in[0,1],\\
f^R_h(r\mid s,a)&\leq& \eta, \quad \forall r\in[0,1].
\end{eqnarray*}
\end{assumption}
Next, we introduce the local Lipschitz property in Lemma \ref{lemma_near_lipschitz}, and then derive a key inequality to convert the regret under the Discretization Environment to the regret under the Original Environment in Lemma \ref{lemma_regret_conversion}. 
\begin{lemma}[Local Lipschitz Property of the Q Function of a Markovian Policy]
\label{lemma_near_lipschitz}
Under Assumption \ref{assumption_lipschitz_continuous_bounded}, given any Markovian policy $\pi\in\Pi$, 
we have the corresponding Q function satisfying for any $(s,a)\in\calS\times\calA$, for any $y,y'\in\calY$ with $|y-y'|\leq 1$,
$$|Q_h^{\pi}(s,y,a)-Q_h^{\pi}(s,y',a)|\leq H\kappa(\lambda+2\eta)|y-y'|.$$
\end{lemma}
\begin{proof}[Proof of Lemma \ref{lemma_near_lipschitz}]
For any $h\in [H]$, $(s,a)\in\calS\times\calA$,  for any $y,y'\in\calY$ with $|y-y'|\leq 1$, 
without loss of generality assume $y\leq y'$. We have
\begin{align}
&|Q_{h}^{\pi}(s,y,a)-Q_{h}^{\pi}(s,{y'},a)|
\nonumber\\
&\leq \E_{s'\sim \prob_{h}^S(\cdot\mid s,a)}\left[ \left|\int_{0}^1V_{h+1}^{\pi}(s',y+r)f^R_{h}(r\mid s,a){\rm d}r-
\int_{0}^1V_{h+1}^{\pi}(s',y'+r)f^R_{h}({\rm d}r\mid s,a){\rm d} r\right| \right]
\label{equation_near_lipschitz_1}\\
&\leq \E_{s'\sim \prob_{h}^S(\cdot\mid s,a)}\left[ \left|\int_{y'-y}^1 V_{h+1}^{\pi}(s',y+r)f^R_{h}(r\mid s,a){\rm d}r-
\int_{0}^{1-y'+y}V_{h+1}^{\pi}(s',y'+r)f^R_{h}(r\mid s,a){\rm d} r\right| \right] \nonumber \\
& +\E_{s'\sim \prob_{h}^S(\cdot\mid s,a)}\left[ \left|
\int_{0}^{y'-y} V_{h+1}^{\pi}(s',y+r)f^R_{h}(r\mid s,a){\rm d} r-\int_{1-y'+y}^1 V_{h+1}^{\pi}(s',y'+r)f^R_{h}(r\mid s,a){\rm d} r
\right| \right].
\label{equation_near_lipschitz_2}
\end{align}
\eqref{equation_near_lipschitz_1} holds by the Bellman equation and \eqref{equation_near_lipschitz_2} holds by triangle inequality.

In \eqref{equation_near_lipschitz_2}, by utilizing $f^R_{h}(\cdot \mid s,a)\leq \eta$ and the fact that $U(0)=0$ and hence $V_{h+1}^{\pi}(s',\cdot)\leq H\kappa$, 
we have
\begin{align}
&\E_{s'\sim \prob_{h}^S(\cdot\mid s,a)}\left[ \left|
\int_{0}^{y'-y} V_{h+1}^{\pi}(s',y+r)f^R_{h}(r\mid s,a){\rm d} r-\int_{1-y'+y}^1 V_{h+1}^{\pi}(s',y'+r)f^R_{h}(r\mid s,a){\rm d} r
\right| \right]\nonumber\\
&{\leq \E_{s'\sim \prob_{h}^S(\cdot\mid s,a)}\left[ \left|
\int_{0}^{y'-y} V_{h+1}^{\pi}(s',y+r)f^R_{h}(r\mid s,a){\rm d} r\right|+\left|\int_{1-y'+y}^1 V_{h+1}^{\pi}(s',y'+r)f^R_{h}(r\mid s,a){\rm d} r
\right| \right]}\nonumber\\
&{\leq \E_{s'\sim \prob_{h}^S(\cdot\mid s,a)}\left[ 
\int_{0}^{y'-y} H\kappa\eta{\rm d} r+\int_{1-y'+y}^1 H\kappa\eta{\rm d} r
 \right]}\nonumber\\
&=2H\kappa\eta|y'-y|.
\label{equation_near_lipschitz_3}
\end{align}

It now suffices to bound the first difference term in \eqref{equation_near_lipschitz_2}. We define $r'=r+y'-y$ and observe that 
\begin{align}
&\E_{s'\sim \prob_{h}^S(\cdot\mid s,a)}\left[ \left|\int_{y'-y}^1 V_{h+1}^{\pi}(s',y+r)f^R_{h}(r\mid s,a){\rm d} r-
\int_{0}^{1-y'+y}V_{h+1}^{\pi}(s',y'+r)f^R_{h}(r\mid s,a){\rm d}r\right| \right]\nonumber\\
&=\E_{s'\sim \prob_{h}^S(\cdot\mid s,a)}\Big[ \Big|\int_{y'-y}^1V_{h+1}^{\pi}(s',y+r)f^R_{h}(r\mid s,a){\rm d} r\nonumber\\
&\qquad\qquad\qquad-\int_{y'-y}^1V_{h+1}^{\pi}(s',y+r')f^R_{h}(r'-y'+y\mid s,a){\rm d}r'\Big| \Big]
\label{equation_near_lipschitz_4}\\
&\leq H \kappa \E_{s'\sim \prob_{h}^S(\cdot\mid s,a)}\left[ 
\int_{y'-y}^1\Big|f^R_{h}(r\mid s,a)-f^R_{h}(r-y'+y\mid s,a)\Big|{\rm d} r
\right]\label{equation_near_lipschitz_5}\\
&\leq H\kappa\lambda |y'-y|.\label{equation_near_lipschitz_6}
\end{align}
\eqref{equation_near_lipschitz_4} holds by change of variables. \eqref{equation_near_lipschitz_5} holds by H\"older's inequality with $p=\infty$ and $q=1$. 
\eqref{equation_near_lipschitz_6} holds by assumption \ref{assumption_lipschitz_continuous_bounded}. 
Finally, combining \eqref{equation_near_lipschitz_2}, \eqref{equation_near_lipschitz_3}, and \eqref{equation_near_lipschitz_6} yields the desired result.
\end{proof}

\begin{remark}[Remark on Lemma \ref{lemma_near_lipschitz}]
Lemma \ref{lemma_near_lipschitz} shows that the Q function is locally Lipschitz for $|y-y'|\leq 1$. Therefore, for any Markovian policy $\pi$, at any timestamp $h\in [H]$, for any $y,y'\in\calY_h$ with $y\leq y'$, we have the following decomposition: 
\begin{align}
&|Q_h^{\pi}(s,y)-Q_h^{\pi}(s,y')|\nonumber\\
&\qquad \leq 
\sum_{i=0}^{\lfloor{y'-y}\rfloor-1}|Q_h^{\pi}(s,y+i)-Q_h^{\pi}(s,y+i+1)|+|Q_h^{\pi}(s,y+\lfloor{y'-y}\rfloor)-Q_h^{\pi}(s,y')|.
\end{align}
Due to the local Lipschitz property, each difference term can be bounded, and therefore we can show that $Q_h^{\pi}$ is actually Lipschitz continuous as $\mathcal{Y}_h=[0,h]$ is a bounded set. 
\end{remark}
Next, we show that the regret under the Discretized Environment can be converted to the regret under the Original Environment. 

\begin{lemma}[Regret Conversion] 
\label{lemma_regret_conversion}
Under Assumption \ref{assumption_lipschitz_continuous_bounded},
if we define the regret under the Discretized Environment as
\begin{equation*}
\overline{{\rm Regret}}(K):=\sum_{k=1}^{K} \Big(\barV_1^{{\barpi^*}}(s_1^k,0) - \barV_1^{\pi^k}(s_1^k,0)\Big),
\end{equation*}
then we can convert it to a high probability regret bound under the original MDP. That is, with probability at least $1-\frac{p}{2}$, 
\begin{equation*}
\sum_{k=1}^{K} \Big(V_1^{\pi^*}(s_1^k,0) - V_1^{\Gamma(\pi^k)}(s_1^k,0)\Big) \leq \overline{{\rm Regret}}(K)+ \sqrt{H^2T \kappa^2 {\iota_2}} + \frac{1}{2}HT\kappa(\lambda+2\eta)\epsilon_o+T\kappa\epsilon_o,
\end{equation*}
{\color{black}where $\iota_2:={\rm log} \left(\frac{p\epsilon_o}{16H^2SAK}\right)$.}
\end{lemma}
\begin{proof}[Proof of Lemma \ref{lemma_regret_conversion}]
We have the following decomposition:

\begin{align}
&\sum_{k=1}^{K} \left(V_1^{\pi^*}(s_1^k,0) - V_1^{\Gamma(\pi^k)}(s_1^k,0)\right) \nonumber\\
&= \sum_{k=1}^{K} \left(V_1^{\pi^*}(s_1^k,0) - \barV_1^{{\barpi^*}}(s_1^k,0) \right)
+ \sum_{k=1}^{K} \left( \barV_1^{{\barpi^*}}(s_1^k,0) - \barV_1^{\pi^k}(s_1^k,0) \right)\nonumber\\
&\qquad\qquad\qquad\qquad+\sum_{k=1}^{K} \left(\barV_1^{\pi^k}(s_1^k,0)  -V_1^{\Gamma(\pi^k)}(s_1^k,0) \right)\nonumber\\
&\leq \sum_{k=1}^{K}\Big(\barV_1^{\pi^k}(s_1^k,0) -V_1^{\Gamma(\pi^k)}(s_1^k,0)\Big)+\overline{{\rm Regret}}(K)+ T\kappa\epsilon_o.\label{equation_regret_conversion_3}
\end{align}
\eqref{equation_regret_conversion_3} holds due to Proposition \ref{lemma:v_difference_ERSMDP_BARMDP}.
It suffices to bound the first summation term from \eqref{equation_regret_conversion_3}, which is the summation of performance difference of $\pi^k$ under different environments. Observe that 
\begin{align}
&\barV_h^{\pi^k}(s_h^k,y_h^k) - V_h^{\Gamma(\pi^k)}(s_h^k,y_h^k)\nonumber\\
&=\barprob_h \barV_{h+1}^{\pi^k}(s_h^k,y_h^k,a_h^k)-\prob_h V_{h+1}^{\Gamma(\pi^k)}(s_h^k,y_h^k,a_h^k)\nonumber\\
&=\barprob_h (\barV_{h+1}^{\pi^k}-V_{h+1}^{\Gamma(\pi^k)})(s_h^k,y_h^k,a_h^k)
+(\barprob_h-\prob_h) V_{h+1}^{\Gamma(\pi^k)}(s_h^k,y_h^k,a_h^k)\nonumber\\
&=\barV_{h+1}^{\pi^k}(s_{h+1}^k,y_{h+1}^k)-V_{h+1}^{\Gamma(\pi^k)}(s_{h+1}^k,y_{h+1}^k)+\sigma_h^k+(\barprob_h-\prob_h) V_{h+1}^{\Gamma(\pi^k)}(s_h^k,y_h^k,a_h^k),\label{equation_regret_conversion_4}
\end{align}
where $\sigma_ h^k:=\barprob_h (\barV_{h+1}^{\pi^k}-V_{h+1}^{\Gamma(\pi^k)})(s_h^k,y_h^k,a_h^k)-(\barV_{h+1}^{\pi^k}-V_{h+1}^{\Gamma(\pi^k)})(s_{h+1}^k,y_{h+1}^k)$. In addition, recall that $\barr=\projR{r}$, and we have
\begin{align}
&(\barprob_h-\prob_h) V_{h+1}^{\Gamma(\pi^k)}(s_h^k,y_h^k,a_h^k)
=\nonumber\\
&\qquad\E_{s'\sim \prob_h^S(\cdot\mid s_h^k,a_h^k)}\left[ \int_{0}^1\left(V_{h+1}^{\Gamma(\pi^k)}(s^\prime,y_h^k+\barr)-
V_{h+1}^{\Gamma(\pi^k)}(s^\prime,y_h^k+r)\right)R_h({\rm d}r\mid s_h^k,a_h^k) \right].\label{equation_regret_conversion_4.5}
\end{align}
By the construction of $\pi^k$, 
$$\pi_{h+1}^k(s^\prime,y_h^k+r)=\pi_{h+1}^k(s^\prime,y_h^k+\barr).$$
Let $a':=\pi_{h+1}^k(s^\prime,y_h^k+\barr)$. It holds that
\begin{align}
&V_{h+1}^{\Gamma(\pi^k)}(s^\prime,y_h^k+\barr)-
V_{h+1}^{\Gamma(\pi^k)}(s^\prime,y_h^k+r) \nonumber\\
&=Q_{h+1}^{\Gamma(\pi^k)}(s^\prime,y_h^k+\barr,a')-
Q_{h+1}^{\Gamma(\pi^k)}(s^\prime,y_h^k+r,a') \nonumber\\
&\leq \frac{1}{2}H\kappa(\lambda+2\eta)\epsilon_o.\label{equation_regret_conversion_5}
\end{align}
\eqref{equation_regret_conversion_5} holds by Lemma \ref{lemma_near_lipschitz}.
Combining \eqref{equation_regret_conversion_4}, \eqref{equation_regret_conversion_4.5}, and \eqref{equation_regret_conversion_5} yields
\begin{align}
&\barV_h^{\pi^k}(s_h^k,y_h^k) - V_h^{\Gamma(\pi^k)}(s_h^k,y_h^k)\nonumber\\
&\leq \barV_{h+1}^{\pi^k}(s_{h+1}^k,y_{h+1}^k)-V_{h+1}^{\Gamma(\pi^k)}(s_{h+1}^k,y_{h+1}^k)+\sigma_h^k+\frac{1}{2}H\kappa(\lambda+2\eta)\epsilon_o.\label{equation_regret_conversion_6}
\end{align}
Applying \eqref{equation_regret_conversion_6} recursively gives
\begin{align}
&\sum_{k=1}^{K}(\barV_1^{\pi^k}(s_1^k,0) -V_1^{\Gamma(\pi^k)}(s_1^k,0))\nonumber\\
&\leq \sum_{k=1}^{K}(\barV_{H+1}^{\pi^k}(s_{H+1}^k,y_{H+1}^k)-V_{H+1}^{\Gamma(\pi^k)}(s_{H+1}^k,y_{H+1}^k)) + \sum_{k=1}^{K}\sum_{h=1}^{H} \sigma_h^k + \frac{1}{2}H^2K\kappa(\lambda+2\eta)\epsilon_o\nonumber\\
&=\sum_{k=1}^{K}\sum_{h=1}^{H} \sigma_h^k + \frac{1}{2}HT\kappa(\lambda+2\eta)\epsilon_o.\label{equation_regret_conversion_7}
\end{align}
The last equality holds because 
$$\barV_{H+1}^{\pi^k}(s_{H+1}^k,y_{H+1}^k)=V_{H+1}^{\Gamma(\pi^k)}(s_{H+1}^k,y_{H+1}^k)=U(y_{H+1}^k).$$
Next, note that $\sigma_h^k$ is a martingale difference sequence. By applying the Azuma-Hoeffding inequality, we have with probability at least $1-\frac{p}{2}$,
\begin{align}
\sum_{k=1}^{K}\sum_{h=1}^{H} \sigma_h^k \leq \sqrt{H^2T \kappa^2 {\iota_2}}.\label{equation_regret_conversion_8}
\end{align}
Combining \eqref{equation_regret_conversion_3}, \eqref{equation_regret_conversion_7}, and \eqref{equation_regret_conversion_8} finishes the proof.
\end{proof} 

With Lemma \ref{lemma_regret_conversion} in hand,  it suffices to develop an algorithm with bounded $\overline{\rm Regret}(K)$ in order to achieve bounded regret under the Original Environment.

\subsection{Value Iteration for General Utility Function with Upper Confidence Bound}\label{sec:VIGU-UCB}

To address the challenge of RL under the general utility function when the simulator is not available, we develop a value iteration method, denoted VIGU-UCB, based on the upper confidence bound {using episodes of complete trajectories from timestamp 0 to timestamp H}. 

The details are described in Algorithm \ref{alg:VICNUCB}. At the start of each episode (lines 2-7), we first estimate the transition kernel $\prob^S$ and the reward distribution $R_h$, with $\prob^S$ estimated by counting numbers and $R_h$ estimated by counting numbers with a projection. The resulting $\hatR_h$ is an unbiased estimator of the discrete approximation $\barR_h$, as discussed in Section \ref{subsection_proof_theorem_3}. In the next step (lines 8-20)  we compute a Hoeffding-type high probability upper bound of the Q function based on the estimated transition kernels and take the corresponding greedy action.                                                                                

\begin{algorithm}
\caption{Value Iteration under General Utility with Upper Confidence Bond (VIGU-UCB)}
\label{alg:VICNUCB}
\begin{algorithmic}[1]
\State Initialize: $\forall (s,y,a) \in \calS \times \barY_{H+1} \times \calA,\ \hat{Q}_{H+1}(s,y,a)= U(y)$. $\forall h \in [H],\ \mathcal{D}_h=\emptyset,\ \mathcal{C}_h=\emptyset.\ \forall k \in [K],\ y_1^k=0.$
\For{$k=1,2,\cdots K$}
\For{$h= H-1,\cdots 1$}
\For{$(s,a) \in \calS \times \calA $}
\If{$(s,a) \in \mathcal{C}_h$}
\State Estimate the transition $\hat{\prob}^S_h(s^\prime \mid s,a)$ for all $s^\prime \in \calS$: 
\begin{align*}
&N_h^k(s,a,s')=\sum_{(s_h,a_h,r_h,s_{h+1})\in \calD_h}\mathbbm{1}\{(s_h,a_h,s_{h+1})=(s,a,s')\},\\
&N_h^k(s,a)=\sum_{s'\in\calS}N_h^k(s,a,s'),\\
&\hat{\prob}^S_h(s^\prime \mid s,a)=\frac{N_h^k(s,a,s')}{N_h^k(s,a)}.
\end{align*}
\State Estimate the transition $\hatR_h(r \mid s,a)$ for all $r\in \barI$ defined in Section \eqref{definition_rtilde}: 
$$\hatR_h(r \mid s,a)=\frac{\sum_{(s_h,a_h,r_h,s_{h+1})\in \calD_h} \mathbbm{1}\{{\rm Proj}_{\barI}(r_h)=r\}}{N_h^k(s,a)}.$$
\For{$y\in\barY_h$}
\State Compute the estimation of the Q function:
\begin{equation*}
\hat{Q}_h(s,y,a) = \min\left\{\hat{\prob}_h \hat{V}_{h+1}(s,\projY{y},a)+b^k_h(s,a),H\kappa \right\},    
\end{equation*}
$\qquad\qquad\qquad\quad$ where $b^k_h(s,a)=\sqrt{\frac{H^2\kappa^2{\iota_2}}{N_h^k(s,a)}}$ with ${\iota_2}={\rm log}(\frac{p\epsilon_o}{16H^2SAK})$, and
$$\hat{\prob}_h(s^\prime,y^\prime \mid s,y,a)=\hat{\prob}_h^S\left(s^\prime \mid s,y,a\right)\hat{R}_h(y^\prime-y \mid s,a).$$
\label{line:p_hat}
\label{line:qhat}
\State Compute $\hat{V}_h(s,y) = \max_{a\in\calA}  \hat{Q}_h(s,y,a)$. 
\label{line:vhat}
\EndFor
\EndIf
\EndFor
\EndFor
\For{$h = 1,2,\dots H$}

\State Take action $a_h^k \in arg\max_{a\in\calA} \hat{Q}_h(s_h^k,y_h^k,a)$.
\label{line:a_k}
\State Observe $s_{h+1}^k,r_h^k$ and update $D_h=D_h \cup (s_h^k,a_h^k,r_h^k,s_{h+1}^k)$, $\mathcal{C}_h = \mathcal{C}_h\cup \{s_h^k,a_h^k\} $.
\State Update $y_{h+1}^k=y_h^k+\projR{r_h^k}$. 
\EndFor
\EndFor
\end{algorithmic}
\end{algorithm}

\subsection{Theoretical Analysis of VIGU-UCB}
\label{sec:4.3}

Under Assumption \ref{assumption_lipschitz_continuous_bounded}, we are able to bound the discretization error incurred by the covering net. The regret is bounded formally in the following theorem.
\begin{theorem}
\label{thm:VIUCB_regret}
Let $\pi^k$ be the greedy policy induced from Line \ref{line:a_k} of Algorithm \ref{alg:VICNUCB} in episode $k$. 
Namely, define $\pi^k:=\{\pi_h^k\}_{h\in[H]}$ with $\pi_h^k(s,y) \in \arg\max_a \hat{Q}_h^k(s,y,a)$ for any $(s,y)\in\calS\times\barY_h$. 
In Algorithm \ref{alg:VICNUCB}, the agent is taking action according to $\Gamma(\pi^k)$. Then we have
under Assumption \ref{assumption_lipschitz_continuous_bounded}, with high probability, the regret of Algorithm \ref{alg:VICNUCB} is bounded by polynomial terms in the total time $T=HK$. That is, with probability at least $1-p$,
$${\rm Regret}(K)=\calO \left(
\sqrt{H^3SAT\kappa^2{\iota_2}}+\sqrt{H^2T\kappa^2{\iota_2}}+\frac{H^3S^2A{\iota_2}^2}{\epsilon_o} 
+ HT\kappa(\lambda+\eta)\epsilon_o
\right),$$ 
{\color{black}where $\iota_2$ is a logarithm term depending on $p$, which is defined in Lemma \ref{lemma_regret_conversion}.}  Moreover, if we pick $\epsilon_o={\Theta}\left(\sqrt{\frac{H^2S^2A}{{T}\kappa(\lambda+\eta)}}\right)$,  we have with probability at least $1-p$, $${\rm Regret}(K)=\calO \left(\sqrt{H^3SAT\kappa^2\iota_3}+\sqrt{H^2T\kappa^2\iota_3}+\sqrt{H^4S^2AT\kappa(\lambda+\eta)\iota_3^2}\right),$$ 
with $\iota_3:={\Theta}\left(\log\sqrt{\frac{p^2}{AT^3\kappa(\lambda+\eta)}}\right)$.
\end{theorem}

{\color{black} Theorem \ref{thm:VIUCB_regret} demonstrates that Algorithm \ref{alg:VICNUCB} efficiently achieves a regret upper bound that is polynomial in the model parameters. The proof of Theorem \ref{thm:VIUCB_regret} is detailed in Section \ref{sec:proof_regret}. Compared to the regret in the risk-neutral setting \citep{azar2017minimax}, Algorithm \ref{alg:VICNUCB} has a regret of the same order in $A$ and $T$, but with a higher order in $H$ and $S$. We believe this discrepancy is fundamental to our setting because of the need to augment the state space with a new state accounting for the cumulative (stochastic) reward, as opposed to the risk-neutral RL where the primary difficulty lies in efficiently estimating the transition kernel with {\it finite} state and action spaces. Estimating transition kernels associated with cumulative reward distribution (which is continuous) is substantially more complex. } 

{\color{black}Technically, in Theorem \ref{thm:VIUCB_regret}, the leading terms regarding $H$, $S$, $A$, and $T$ are} 
\begin{equation*}
{\color{black}\calO \Big(\frac{H^3S^2A{\iota_2}^2}{\epsilon_o} 
\Big)+\calO\Big(HT\kappa(\lambda+\eta)\epsilon_o\Big).}    
\end{equation*}
{\color{black}These two terms trade off the state space cardinality and the approximation error through $\epsilon_o$: the discretization of the additional dimension in the state space has a cardinality of  $\frac{H}{\epsilon_o}$, to which the order of regret scales up. On the other hand, the discretization error is of order $\calO\left(HT\kappa(\lambda+\eta)\epsilon_o\right)$ and scales with $\epsilon_o$.} 

{\color{black} In contrast,  without the need for discretization, the leading term is $\calO \left(\sqrt{H^3SAT\kappa^2{\iota_2}}\right)$ for a Hoeffding-type UCB algorithm in the risk-neutral setting \citep{azar2017minimax}. This result  has a lower order in $H$ and $S$.}


\subsection{Proof of Theorem \ref{thm:VIUCB_regret}}
\label{sec:proof_regret}
With Lemma \ref{lemma_regret_conversion} in hand,  it suffices to bound $\overline{\rm Regret}(K)$ in order to prove Theorem \ref{thm:VIUCB_regret}.  In the next lemma, we show that the value functions calculated from Algorithm \ref{alg:VICNUCB} are high probability upper bounds of the optimal value function under the Discretized Environment. The analysis of the lemma follows the proof from existing literature \citep{azar2017minimax} for the risk-neutral setting, with adaptation to our risk-sensitive settings.

\begin{lemma}[UCB]
\label{lemma:UCB}
Let $\hat{Q}_h^k,\ \hat{V}_h^k$ denote the value functions in Lines \ref{line:qhat} and \ref{line:vhat} at timestamp $h$ in episode $k$. 
Then these functions are a high probability upper bound of the optimal value functions under $\BARMDP$. Mathematically, with probability at least $1-\frac{p}{8}$, for any $h\in[H+1]$ and for any $(k,s,y,a) \in [K]\times \calS \times {\calY_h}\times\calA$, 
\begin{align}
\hat{Q}^k_h(s,y,a)&\geq \barQ_h^{{{\barpi^*}}}(s,y,a), \label{equation_UCB_3}\\
\hat{V}^k_h(s,y)&\geq \barV_h^{{{\barpi^*}}}(s,y).  \label{equation_UCB_4}  
\end{align}
\end{lemma}

\begin{proof}[Proof of Lemma \ref{lemma:UCB}.]
By Hoeffding's inequality, we have with probability at least $1-\frac{p}{8}$, for any $(k,h,s,a)\in[K]\times[H]\times[S]\times[A]$,
\begin{align}
    \left|(\hat{\prob}_h-\prob_h)\barV^{{{\barpi^*}}}_{h+1}(s,y,a)\right|\leq b_h^k(s,a). \label{equation_UCB_2}
\end{align}

Next, we prove the statement by induction. Under the event that \eqref{equation_UCB_2} holds for any $(k,h,s,a)\in[K]\times[H]\times[S]\times[A]$, suppose for any $h'=h+1,h+2\cdots H+1$, for any $ (s,y,a) \in  \calS \times {\calY_{h'}} \times \calA$, it holds that $$\hat{Q}_{h'}(s,y,a)\geq \barQ_{h'}^{{{\barpi^*}}}(s,y,a),$$ $$\hat{V}_{h'}(s,y)\geq \barV_{h'}^{{{\barpi^*}}}(s,y).$$
From Line \ref{line:qhat} in Algorithm $\ref{alg:VICNUCB}$, 
$$\hat{Q}_h^k(s,y,a) = \min\left\{\hat{\prob} \hat{V}_{h+1}^k(s,\projY{y},a)+b^k_h(s,a),H\kappa \right\}.$$
Then we have two cases:

1. $\hat{Q}^k_h(s,y,a) = H\kappa  $. In this case, $$\hat{Q}^k_h(s,y,a) = H\kappa  \geq \barQ^{{\barpi^*}}_h(s,y,a),$$
$$\hat{V}_h^k(s,y)=H\kappa \geq \barV_h^{{{\barpi^*}}}(s,y).$$ 

2. $\hat{Q}^k_h(s,y,a) = \hat{\prob}_h\hat{V}_{h+1}^k(s,y,a)+b_h^k(s,a)$. In this case, we have
\begin{align}
&\hat{Q}^k_h(s,y,a)-\barQ^{{{\barpi^*}}}_h(s,y,a)\nonumber\\
    &=\hat{\prob}_h\hat{V}_{h+1}^k(s,y,a)-\prob_h \barV^{{{\barpi^*}}}_{h+1}(s,y,a)+b_h^k(s,a) \nonumber\\
    &\geq \hat{\prob}_h\barV^{{{\barpi^*}}}_{h+1}(s,y,a)-\prob_h \barV^{{{\barpi^*}}}_{h+1}(s,y,a)+b_h^k(s,a)
    \label{equation_UCB_1}\\
    &\geq -\left|(\hat{\prob}_h-\prob_h)\barV^{{{\barpi^*}}}_{h+1}(s,y,a)\right|+b_h^k(s,a)\nonumber\\
    &\geq 0. \label{equation_UCB_2.5}
\end{align}
\eqref{equation_UCB_1} holds by induction and \eqref{equation_UCB_2.5} holds due to \eqref{equation_UCB_2}. Then we have 
$$\hat{V}_h^k(s,y)=\max_{a\in\calA} \hat{Q}_h^k(s,y,a)\geq \max_{a\in\calA}\barQ_h^{{{\barpi^*}}}(s,y,a)= \barV_h^{{{\barpi^*}}}(s,y).$$ 

Lastly, we can check that at timestamp $h'=H+1$, $\hat{Q}^k_{H+1}(s,y,a)=U(y)=
\barQ^{{{\barpi^*}}}_{H+1}(s,y,a)$ by the algorithm design.
\end{proof}

With this upper confidence property, we show that the regret under the Discretized Environment can be decomposed in the following lemma.
\begin{lemma}[Error Decomposition] \label{lemma:decomp}
Under the Discretized Environment, the difference between the value function calculated from Algorithm \ref{alg:VICNUCB} and the true value function of $\pi^k$ can be upper bounded. Formally, with probability of at least $1-\frac{p}{8}$, for any $(k,h)\in[K]\times [H]$,
\begin{align}
\hat{V}_1^k(s_1^k,0)-\barV^{\pi^k}_1(s_1^k,0)
\leq e\sum_{h=1}^{H}\left(
\left(1+\frac{1}{H}\right)\zeta_{h}^k + \delta_{h}^k + 2b_{h}^k\right),
\end{align}
where 
\begin{align}
\delta_h^k&:=\frac{2H^2S\kappa{\iota_2}}{\epsilon_o N_h^k(s_h^k,a_h^k)} \text{ with } {\iota_2} \text{ defined in Theorem \ref{thm:bound},} \label{delta} \\
\zeta_h^k&:=\barprob_h(\hat{V}_{h+1}^k-\barV^{\pi^k}_{h+1})(s_h^k,y_h^k,a_h^k)- \left(\hat{V}_{h+1}^k(s_{h+1}^k,y_{h+1}^k)-\barV^{\pi^k}_{h+1}(s_{h+1}^k,y_{h+1}^k)\right), \\
b_{h}^k&:=b_h^k(s_h^k,a_h^k) \text{ with } b_h^k(\cdot,\cdot) \text{ defined in Line \ref{line:qhat} of Algorithm \ref{alg:VICNUCB}.}
\end{align}
\end{lemma}
\begin{proof}[Proof of Lemma \ref{lemma:decomp}]
We first derive a recursive format for $\hat{V}_h^k(s_h^k,y_h^k)-\barV^{\pi^k}_h(s_h^k,y_h^k)$. It holds that
\begin{align}
&\hat{V}_h^k(s_h^k,y_h^k)-\barV^{\pi^k}_h(s_h^k,y_h^k)\nonumber\\
&=\hat{Q}_h^k(s_h^k,y_h^k,a_h^k)-\barQ^{\pi^k}_h(s_h^k,y_h^k,a_h^k)\nonumber\\
&\leq \hat{\prob}_h\hat{V}_{h+1}^k(s_h^k,y_h^k,a_h^k)-\barprob_h \barV^{\pi^k}_{h+1}(s_h^k,y_h^k,a_h^k)+b_h^k\label{equ:lemma_error_decomposition_recursion_3}\\
\label{equ:lemma_error_decomposition_recursion_1}
&=(\hat{\prob}_h-\barprob_h)\hat{V}_{h+1}^k(s_h^k,y_h^k,a_h^k)
+\barprob_h(\hat{V}_{h+1}^k-\barV^{\pi^k}_{h+1})(s_h^k,y_h^k,a_h^k)
+b_h^k\\
&=(\hat{\prob}_h-\barprob_h)(\hat{V}_{h+1}^k-\barV^{\pi^k}_{h+1})(s_h^k,y_h^k,a_h^k)
+(\hat{\prob}_h-\barprob_h)\barV^{\pi^k}_{h+1}(s_h^k,y_h^k,a_h^k)\nonumber\\
&\qquad+\barprob_h(\hat{V}_{h+1}^k-\barV^{\pi^k}_{h+1})(s_h^k,y_h^k,a_h^k)
+b_h^k.\label{equ:lemma_error_decomposition_recursion_2}
\end{align}
\eqref{equ:lemma_error_decomposition_recursion_3} holds by the definition of $\hat{Q}_h^k$ and the Bellman equation for $\barQ^{\pi^k}_h$. \eqref{equ:lemma_error_decomposition_recursion_1} and \eqref{equ:lemma_error_decomposition_recursion_2} holds due to telescoping sums. In addition, because $\hat{\prob}_h$ is an unbiased estimator of $\barprob_h$, by Hoeffding's inequality, we have with probability at least $1-\frac{p}{16}$, for any $(k,h)\in[K]\times [H]$,
\begin{align}
(\hat{\prob}_h-\barprob_h)\barV^{\pi^k}_{h+1}(s_h^k,y_h^k,a_h^k)\leq b_h^k.
\label{equ:recursion}
\end{align}
Let $F(s,y):=\hat{V}_{h+1}^k(s,y)-\barV^{\pi^k}_{h+1}(s,y)$. We have
\begin{align}
&(\hat{\prob}_h-\Bar{\prob}_h)(\hat{V}_{h+1}^k-\barV^{\pi^k}_{h+1})(s_h^k,y_h^k,a_h^k)\nonumber\\
&=\sum_{s^\prime\in\calS,r\in\barI}\left(\hat{\prob}_h(s^\prime,y+r\mid s_h^k,y_h^{k},a_h^k)-\Bar{\prob}_h(s^\prime,y+r\mid s_h^k,y_h^{k},a_h^k)\right)F(s^\prime,y+r).
\nonumber
\end{align}
Apply Bernstein's inequality and we have with probability at least $1-\frac{p}{16}$,
\begin{align}
&\sum_{s^\prime\in\calS,r\in\barI}\left(\hat{\prob}_h(s^\prime,y+r\mid s_h^k,y_h^{k},a_h^k)-\Bar{\prob}_h(s^\prime,y+r\mid s_h^k,y_h^{k},a_h^k)\right)F(s^\prime,y+r)\nonumber\\
&\leq \sum_{s^\prime\in\calS,r\in\barI}\left(\sqrt{\frac{\Bar{\prob}_h(s^\prime,y+r\mid s_h^k,y_h^{k},a_h^k))}{N_h^k(s_h^k,a_h^k)}}+\frac{{\iota_2}}{N_h^k(s_h^k,a_h^k)}\right)F(s^\prime,y+r)\label{equ:bern}\\
&= \sum_{s^\prime\in\calS,r\in\barI}\left(\sqrt{\frac{\Bar{\prob}_h(s^\prime,y+r\mid s_h^k,y_h^{k},a_h^k))}{H}\frac{H}{N_h^k(s_h^k,a_h^k)}}+\frac{{\iota_2}}{N_h^k(s_h^k,a_h^k)}\right)F(s^\prime,y+r)\nonumber\\
\label{equ:sqrtab}
&\leq \sum_{s^\prime\in\calS,r\in\barI}({\frac{\Bar{\prob}_h(s^\prime,y+r\mid s_h^k,y_h^{k},a_h^k))}{H}}+\frac{H+{\iota_2}}{N_h^k(s_h^k,a_h^k)})F(s^\prime,y+r)\\
&\leq \frac{1}{H}{\Bar{\prob}_h}(\hat{V}_{h+1}^k-\barV^{\pi^k}_{h+1})(s_h^k,y_h^k,a_h^k)+\delta_h^k,
\label{equ:bernresult}
\end{align}
where $\delta_h^k$ is defined in \eqref{delta}. 
\eqref{equ:bern} follows by Bernstein's inequality.
\eqref{equ:sqrtab} holds using the fact that $\forall a,b>0$, it holds that $\sqrt{ab}\leq a+b$.
\eqref{equ:bernresult} holds because $F(s^\prime,y+r)\leq H\kappa$.

By combining \eqref{equ:lemma_error_decomposition_recursion_2},  \eqref{equ:recursion}, and \eqref{equ:bernresult} we have with probability at least $1-\frac{p}{8}$, for any $(h,k)\in[H]\times[K]$,
\begin{align}
&\hat{V}_h^k(s_h^k,y_h^k)-\barV^{\pi^k}_h(s_h^k,y_h^k)\nonumber\\
&\leq (\hat{\prob}_h-\barprob_h)(\hat{V}_{h+1}^k-\barV^{\pi^k}_{h+1})(s_h^k,y_h^k,a_h^k)
+\barprob_h(\hat{V}_{h+1}^k-\barV^{\pi^k}_{h+1})(s_h^k,y_h^k,a_h^k)
+2b_h^k
\nonumber\\
&\leq (1+\frac{1}{H})\barprob_h(\hat{V}_{h+1}^k-\barV^{\pi^k}_{h+1})(s_h^k,y_h^k,a_h^k) 
+\delta_h^k
+2b_h^k\nonumber\\
&=(1+\frac{1}{H})\left(\hat{V}_{h+1}^k(s_{h+1}^k,y_{h+1}^k)-\barV^{\pi^k}_{h+1}(s_{h+1}^k,y_{h+1}^k)\right)
+(1+\frac{1}{H})\zeta_h^k + \delta_h^k + 2b_h^k.
\label{equ:hatvminusv}
\end{align}

Apply \eqref{equ:hatvminusv} recursively and we have with probability at least $1-\frac{p}{8}$,
\begin{align}
&\hat{V}_1^k(s_1^k,0)-\barV^{\pi^k}_1(s_1^k,0)\nonumber\\
&=\sum_{h=1}^{H}(1+\frac{1}{H})^{H-h+1}\left(
(1+\frac{1}{H})\zeta_{h}^k + \delta_{h}^k + 2b_{h}^k\right)+\hat{V}_{H+1}^k(s_{H+1}^k,y_{H+1}^k)-\barV^{\pi^k}_{H+1}(s_{H+1}^k,y_{H+1}^k)\nonumber\\
&\leq e\sum_{h=1}^{H}\left(
(1+\frac{1}{H})\zeta_{h}^k + \delta_{h}^k + 2b_{h}^k\right). \label{equation_regret_decomposition_10}
\end{align}
The last step holds because $(1+\frac{1}{H})^H\leq e$ and $\hat{V}_{H+1}^k(s_{H+1}^k,y_{H+1}^k)=\barV^{\pi^k}_{H+1}(s_{H+1}^k,y_{H+1}^k)=U(y_{H+1}^k)$.
\end{proof}

Now we are ready to bound the regret under the Discretized Environment. 

\begin{lemma}[Regret Bound under the Discretized Environment]
\label{lemma_regret_bound_discretized}
With probability at least $1-\frac{p}{2}$,  the regret under the Discretized Environment can be upper-bounded as the following:
\begin{align}
\sum_{k=1}^K \barV_1^{{\barpi^*}}(s_1^k,0)-\barV_1^{\pi^k}(s_1^k,0) =\calO \left(\sqrt{H^3SAT\kappa^2{\iota_2}}+\sqrt{H^2T\kappa^2{\iota_2}}+\frac{4H^3S^2A{\iota_2}^2}{\epsilon_o}\right).  
\end{align}

\end{lemma}

\begin{proof}[Proof of Lemma \ref{lemma_regret_bound_discretized}] 
By \eqref{equation_regret_decomposition_10}, we have with probability at least $1-\frac{p}{4}$,

\begin{align}
\sum_{k=1}^K \barV_1^{{\barpi^*}}(s_1^k,0)-\barV_1^{\pi^k}(s_1^k,0)  &\leq \sum_{k=1}^K \hat{V_1}^{k}(s_1^k,0)-\barV_1^{\pi^k}(s_1^k,0)\label{equation_regret_under_discretized_environment_1}\\
&\leq e\sum_{k=1}^K\sum_{h=1}^{H}\left(
(1+\frac{1}{H})\zeta_{h}^k + \delta_{h}^k + 2b_{h}^k\right).\label{equation_regret_under_discretized_environment_3}
\end{align}   
\eqref{equation_regret_under_discretized_environment_1} is by Lemma \ref{lemma:UCB} and
\eqref{equation_regret_under_discretized_environment_3} is by Lemma \ref{lemma:decomp}. Next, we can bound the summation terms separately.

$\zeta_h^k$ is a martingale difference sequence. By Azuma-Hoeffding inequality, with probability at least $1-\frac{p}{4}$, it holds that
\begin{align*}
\sum_{k=1}^K\sum_{h=1}^H \zeta_h^k\leq \sqrt{H^2T\kappa^2{\iota_2}}.
\end{align*}
To bound the second summation term, we have
\begin{align}
\sum_{k=1}^K\sum_{h=1}^H \delta_h^k&=\frac{2H^2S{\iota_2}}{\epsilon_o} \sum_{k=1}^K\sum_{h=1}^H \frac{1}{N_h^k(s_h^k,a_h^k)}\nonumber\\
&=\frac{2H^2S{\iota_2}}{\epsilon_o}\sum_{(h,s,a)}\sum_{i=1}^{N_h^K(s,a)}\frac{1}{i}\label{equ:theorem_bound_VIUCB_sum_delta_1}\\
&\leq \frac{4H^2S{\iota_2}}{\epsilon_o}\sum_{(h,s,a)}\log\left(N_h^K(s,a)\right)\label{equ:theorem_bound_VIUCB_sum_delta_2}\\
&\leq \frac{4H^3S^2A{\iota_2}^2}{\epsilon_o}.\label{equ:deltabound}
\end{align}
\eqref{equ:theorem_bound_VIUCB_sum_delta_1} holds by changing the summation over $(k,h)$ to count on $(h,s,a)$ separately.
\eqref{equ:theorem_bound_VIUCB_sum_delta_2} holds because $\sum_{i=1}^N\frac{1}{i}\leq 2\log(N)$. 
\eqref{equ:deltabound} holds using  $\log\left(N_h^K(s,a)\right)\leq \log(K)$.

Next, $\sum_{k=1}^K\sum_{h=1}^H b_h^k$ is bounded by pigeonhole principle:
\begin{align}
\sum_{k=1}^K\sum_{h=1}^H b_h^k 
&= \sum_{k=1}^K\sum_{h=1}^H \left( \sqrt{\frac{H^2\kappa^2{\iota_2}}{N_h^k(s_h^k,a_h^k)}}\right)\nonumber\\
&\leq \sqrt{H^2\kappa^2{\iota_2}}\sum_{(h,s,a)}\sum_{i=1}^{N_h^K(s,a)}\sqrt{\frac{1}{i}}\nonumber\\
&\leq 2\sqrt{H^2\kappa^2{\iota_2}}\sum_{(h,s,a)}\sqrt{N_h^K(s,a)}\label{equ:theorem_bound_VIUCB_sum_b_2}\\
&\leq 2\sqrt{H^2\kappa^2{\iota_2}}\sqrt{HSAT}\label{equ:theorem_bound_VIUCB_sum_b_3}\\
&=2\sqrt{H^3SAT\kappa^2{\iota_2}}.\nonumber
\end{align}
\eqref{equ:theorem_bound_VIUCB_sum_b_2} holds because $\sum_{i=1}^N\sqrt{\frac{1}{i}}\leq 2\sqrt{N}$ and 
\eqref{equ:theorem_bound_VIUCB_sum_b_3} holds by the Cauchy-Schwarz inequality: 
$$\sum_{(h,s,a)}\sqrt{N_h^K(s,a)}\leq \sqrt{\sum_{(h,s,a)}N_h^K(s,a)}\sqrt{\sum_{(h,s,a)}1}=\sqrt{HK}\sqrt{HSA}=\sqrt{HSAT}.$$

Adding everything together, we have with probability at least $1-\frac{p}{2}$, for any $(k,h)\in[K]\times [H]$,
$$\sum_{k=1}^K \barV_h^{{\barpi^*}}(s_h^k,y_h^k)-\barV_h^{\pi^k}(s_h^k,y_h^k) = \calO \left(\sqrt{H^3SAT\kappa^2{\iota_2}}+\sqrt{H^2T\kappa^2{\iota_2}}+\frac{4H^3S^2A{\iota_2}^2}{\epsilon_o}\right).$$
\end{proof}

\begin{proof}[Proof of Theorem \ref{thm:VIUCB_regret}]

Combining Lemma \ref{lemma_regret_bound_discretized} and 
Lemma \ref{lemma_regret_conversion} finishes the proof. 
\end{proof}

{\color{black}
\section{Regret Lower Bound}
\label{sec:regret_LB}

In this section, we provide a regret lower bound for our risk-sensitive MDP setting under general utilities. 
To start, we define the risk-sensitive multi-armed bandit (RS-MAB) problem under general utilities as a special case of the RS-MDP with $S=H=1$ and then extend the result to general RS-MDPs. 

\subsection{Regret Lower Bound for Risk-sensitive MAB}

To begin with, define a RS-MAB problem as  ${\texttt{RS-MDP}(\calS, \calA, H, \prob^S,R,U)}$ with $S=H=1$. Let $\classRSMAB$ denote the set of all RS-MABs that satisfy Assumption \ref{assumption_lipschitz_continuous_bounded}, while let $\classRSMDP$ denote the set of all RS-MDPs that satisfy Assumption \ref{assumption_lipschitz_continuous_bounded}.  
Recall that by the definition of ${\texttt{RS-MDP}}$ from Section \ref{sec:setting_RSMDP}, the reward is bounded in $[0,1]$, and the utility function $U$ is assumed to be continuous and strictly increasing. 

We now introduce deterministic algorithms, randomized algorithms, and their relationships. 
 For any $t\in\mathbb{N}^+$, we define  $\mathcal{O}_t$ as the history of observations up to time $t$.\footnote{Note that this ``history" contains all the past information collected by running the algorithm (across episodes) and is different from the notion of ``history" used in Section \ref{sec:setting_RSMDP}, which is within a given episode.} 
 We define a deterministic algorithm $\algd$ as a sequence of mappings from the {\it history of observations} to an action. That is, $\algd:=\{\algd_t\}_{t=0}^\infty$ with $\algd_t: \mathcal{O}_t\rightarrow \calA$.
Let $\algdetclass$ be the class of all deterministic algorithms. In addition, we define a randomized algorithm $\algr\in\mathcal{P}(\algdetclass)$ as a distribution over the deterministic algorithm set. Consequently, $\mathcal{P}(\algdetclass)$ is the class of all randomized algorithms.

The goal here is to derive the regret lower bound of the RS-MAB problem. Note that the randomness in the regret arises from two key sources: the inherent stochasticity of the RS-MAB environment and the randomness introduced by the algorithm itself.  To proceed, we first state an additional assumption.

\begin{assumption}
\label{assumption_derivative_LB} Assume that $U$ is differentiable on $[0,H]$. In addition, assume that there exists $\underline{\kappa}\in\R^+$, so that  $U'(y)\geq \underline{\kappa}$ for any $y\in[0,T]$.
\end{assumption}
Assumption \ref{assumption_derivative_LB} states that the derivative of the utility function is lower-bounded by a positive constant, which is a natural condition for a strictly increasing utility function. Since $U(\cdot)$ is $\kappa$-Lipschitz, it is clear that $\underline{\kappa} <\kappa$.

\begin{theorem}[Regret lower bound]
\label{thm:RSMAB_lower_bound}
Under Assumption \ref{assumption_derivative_LB}, we have for any $T> A$, 
\begin{align}
\inf_{\algr\in\mathcal{P}(\algdetclass)}\sup_{\mathbb{M}\in\classRSMAB}\E_{\algd\sim\algr}\E_{\mathbb{M}}^{\algd}\big[{\rm regret}(T)\big]=\Omega\Big(\sqrt{AT\underline{\kappa}^2}\Big),   
\end{align}
where $\E_{\mathbb{M}}^{\algd}$ denotes the expectation following algorithm $\algd$ under the RS-MAB environment $\mathbb{M}$, and ${\rm regret}(T)$ denotes the regret for a total round of $T$. 
\end{theorem}
 Theorem \ref{thm:RSMAB_lower_bound} states that for any randomized algorithm $\algr\in\algrandclass$, 
the supremum of the expected regret of $\algr$ over 
$\classRSMAB$ is always larger than the regret lower bound. This indicates that we cannot find an algorithm that has a lower regret guarantee than the regret lower bound.

Following the argument in \citet{auer2002nonstochastic}, it is sufficient to consider deterministic
strategies for taking actions, as any randomized algorithm can be written as a distribution over the deterministic algorithms. Mathematically,
\begin{align}
&\inf_{\algr\in\algrandclass}\sup_{\mathbb{M}\in\classRSMAB}\E_{\algd\sim\algr}\E_{\mathbb{M}}^{\algd}\big[{\rm regret}(T)\big]\nonumber\\
&\qquad\qquad\geq \inf_{\algd\in\algdetclass}\sup_{\mathbb{M}\in\classRSMAB}\E_{\mathbb{M}}^{\algd}\big[{\rm regret}(T)\big].\label{eqn:LB_-3.3}
\end{align}
 Thus we will focus on the lower bound for $\algdetclass$.

To prove Theorem \ref{thm:RSMAB_lower_bound}, we construct a set of RS-MABs with $\Xi'_{\rm RS-MAB}:=\{\mathbb{M}^i\}_{i\in[A]}$, where $\mathbb{M}^i\in\classRSMAB$ is defined as an RS-MAB problem with the $i$-th arm being the best arm, and all other arms are identically sub-optimal. That is, $R(\cdot|s_0,a_j)={\rm Bernoulli}(1/2)$ is a Bernoulli distribution with probability $1/2$ taking value $1$ and probability $1/2$ taking value $0$ for any $j\neq i$. Similarly, $R(\cdot|s_0,a_i)={\rm Bernoulli}(1/2+\epsilon)$ is the the Bernoulli distribution with probability $1/2+\epsilon$ taking value $1$ and probability $1/2-\epsilon$ taking value $0$.  With a random index $I\sim {\rm unif\{1,2,\cdots,A\}}$ where ${\rm unif\{1,2,\cdots,A\}}$ is the discrete uniform distribution over $\{1,2,\dots,A\}$, $\mathbb{M}^I$ is uniformly distributed over $\Xi'_{\rm RS-MAB}$. 
Note that for any $\algr\in\mathcal{P}(\algdetclass)$ and $\mathbb{M}\in\classRSMAB$, we have $0\leq\E_{\mathbb{M}}^{\algd}\big[{\rm regret}(T)\big]\leq T$, which implies that the infimum and the supremum both exist. Then we have, for any $T\in\mathbb{N}^+$,

\begin{align}
&\inf_{\algr\in\algrandclass}\sup_{\mathbb{M}\in\classRSMAB}\E_{\algd\sim\algr}\E_{\mathbb{M}}^{\algd}\big[{\rm regret}(T)\big]\nonumber\\
&\qquad\qquad\geq \inf_{\algd\in\algdetclass}\sup_{\mathbb{M}\in\classRSMAB}\E_{\mathbb{M}}^{\algd}\big[{\rm regret}(T)\big]\label{eqn:LB_-3.2}\\
&\qquad\qquad\geq\inf_{\algd\in\algdetclass}\sup_{\mathbb{M}\in\Xi'_{\rm RS-MAB}}\E_{\mathbb{M}}^{\algd}\big[{\rm regret}(T)\big]\label{eqn:LB_-3}\\
&\qquad\qquad\geq\inf_{\algd\in\algdetclass}\E_{I\sim {\rm unif\{1,\cdots,A\}}}\E_{\mathbb{M}^I}^{\algd}\big[{\rm regret}(T)\big].\label{eqn:LB_0}
\end{align}
{\eqref{eqn:LB_-3.2} follows from \eqref{eqn:LB_-3.3}; \eqref{eqn:LB_-3} holds since $\Xi'_{\rm RS-MAB}\subseteq \algdetclass$; and \eqref{eqn:LB_0} holds because the supremum over a set is no less than the expectation over that set.} 

Next, by the definition of regret, we have
\begin{align}
\E_{I\sim {\rm unif\{1,\cdots,A\}}}\E_{\mathbb{M}^I}^{\algd}\big[{\rm regret}(T)\big]
&= \E_{I\sim {\rm unif\{1,\cdots,A\}}}\Big(\E_{\mathbb{M}^I}^{\pi^*(\mathbb{M}^I)}\Big[ U(y_T)\Big]-\E_{\mathbb{M}^I}^{\algd}\Big[ U(y_T)\Big]\Big)\label{eqn:LB_1},
\end{align}
where $\pi^*(\mathbb{M}^I)$ denotes the optimal policy for the RS-MAB problem $\mathbb{M}^I$ with utility function $U$. 

Compared to the risk-neutral setting, the regret lower bound for the risk-sensitive setting is challenging to establish because \eqref{eqn:LB_1} cannot be {\it directly decomposed} since the utility function and the expectation are not interchangeable. We have the following lemma to facilitate the regret decomposition.
\begin{lemma}
\label{lemma_LB_decomposition}
Let $N_I$ denote the number of times action $a_I$ is taken. 
Under Assumption \ref{assumption_derivative_LB}, we have for any $\algd\in\algdetclass$,
\begin{align*}
\E_{\mathbb{M}^I}^{\pi^*(\mathbb{M}^I)}\Big[ U(y_T)\Big]-\E_{\mathbb{M}^I}^{\algd}\Big[ U(y_T)\Big]
\geq \E_{\mathbb{M}^I}^{\algd}\Big[T-N_I\Big]\epsilon\underline{\kappa}.
\end{align*}    
\end{lemma}
\begin{proof}[Proof of Lemma \ref{lemma_LB_decomposition}]
Let $Y$ be a random variable with a bounded support. 
Let $R_0\sim$Bernoulli$(1/2)$, and $R_1\sim$Bernoulli$(1/2+\epsilon)$, both being independent of $Y$.  Then we have 
\begin{align}
&\E[U(Y+R_1)] - \E[U(Y+R_0)] \nonumber\\
&\qquad\qquad= (\frac{1}{2}+\epsilon) \E[U(Y+1)] + (\frac{1}{2}-\epsilon) \E[U(Y)] - \frac{1}{2} \E[U(Y+1)] -\frac{1}{2} \E[U(Y)]\nonumber\\
&\qquad\qquad=\epsilon \E[U(Y+1)]-\epsilon \E[U(Y)]\nonumber\\
&\qquad\qquad\geq \epsilon\underline{\kappa},\label{equation_LB_decomposition_2}
\end{align}
where the last inequality holds due to Assumption
\ref{assumption_derivative_LB}.

Next, let $\mu_{\mathbb{M}^I}^{\algd}$ denote the probability distribution of $y_T$ following algorithm ${\algd}$ under the RS-MAB ${\mathbb{M}^I}$, and let $\mu_{\mathbb{M}^I}^{\pi^*(\mathbb{M}^I)}$ denote the probability distribution of $y_T$ following policy ${\pi^*(\mathbb{M}^I)}$ under the RS-MAB environment ${\mathbb{M}^I}$. 
Observe that under $\mu_{\mathbb{M}^I}^{\pi^*(\mathbb{M}^I)}$, $y_T$ is the summation of $T$ independent Bernoulli random variables following the distribution Bernoulli$(\frac{1}{2}+\epsilon)$, while under $\mu_{\mathbb{M}^I}^{\algd}$, $y_T$ is the summation of $T-N_I$ independent Bernoulli random variables following the distribution Bernoulli$(1/2+\epsilon)$ and $N_I$ independent Bernoulli random variables following the distribution Bernoulli$(1/2)$. Applying \eqref{equation_LB_decomposition_2} recursively finishes the proof.
\end{proof}

\begin{lemma}
\label{lemma_LB_NI}
For any deterministic algorithm $\algd\in\algdetclass$, for any $T\in\mathbb{N}^+$, 
\begin{align*}
\sum_{I\in[A]}\E_{\mathbb{M}^I}^{\algd}\Big[N_I\Big]\leq T+ T\sqrt{\frac{1}{4}A\,T\log\frac{1}{1-4\epsilon^2}}.
\end{align*}
\end{lemma}
Lemma \ref{lemma_LB_NI} is a direct consequence of the argument before equation (36) in \citet{auer2008near}.
Now we are ready to prove Theorem \ref{thm:RSMAB_lower_bound}.
\begin{proof}[Proof of Theorem \ref{thm:RSMAB_lower_bound}.]
Combining \eqref{eqn:LB_0},   \eqref{eqn:LB_1}, and applying Lemma \ref{lemma_LB_decomposition}, we have under Assumption \ref{assumption_derivative_LB}, 
\begin{align}
&\inf_{\algr\in\algrandclass}\sup_{\mathbb{M}\in\classRSMAB}\E_{\algd\sim\algr}\E_{\mathbb{M}}^{\algd}\big[{\rm regret}(T)\big]\nonumber\\
&\qquad\qquad\geq\inf_{\algd\in\algdetclass}\E_{I\sim {\rm unif\{1,\cdots,A\}}}\E_{\mathbb{M}^I}^{\algd}\big[{\rm regret}(T)\big]\label{eqn:LB_6}\\
&\qquad\qquad\geq\inf_{\algd\in\algdetclass}\E_{I\sim {\rm unif\{1,\cdots,A\}}}\E_{\mathbb{M}^I}^{\algd}\Big[T-N_I\Big]\epsilon\underline{\kappa}\label{eqn:LB_7}\\
&\qquad\qquad=\inf_{\algd\in\algdetclass}\frac{1}{A}\sum_{I\in\{1,2,\cdots,A\}}\E_{\mathbb{M}^I}^{\algd}\Big[T-N_I\Big]\epsilon\underline{\kappa}.\label{eqn:LB_8}
\end{align}
\eqref{eqn:LB_6} holds because the supremum over a set is no less than the expectation over that set; \eqref{eqn:LB_7} holds by Lemma \ref{lemma_LB_decomposition}; \eqref{eqn:LB_8} holds by the definition of the expectation.

Apply Lemma \ref{lemma_LB_NI} to the above inequality, and we have
\begin{align}
&\inf_{\algr\in\algrandclass}\sup_{\mathbb{M}\in\classRSMAB}\E_{\algd\sim\algr}\E_{\mathbb{M}}^{\algd}\big[{\rm regret}(T)\big]\nonumber\\
&\qquad\qquad\geq\inf_{\algd\in\algdetclass}\frac{1}{A}\sum_{I\in[A]}\E_{\mathbb{M}^I}^{\algd}\Big[T-N_I\Big]\epsilon\underline{\kappa}\nonumber\\
&\qquad\qquad\geq   \inf_{\algd\in\algdetclass}
T\epsilon\underline{\kappa} - \frac{1}{A}\epsilon\underline{\kappa}\Big(T+ T\sqrt{\frac{1}{4}AT\log\frac{1}{1-4\epsilon^2}}\Big)\nonumber\\
&\qquad\qquad\geq \inf_{\algd\in\algdetclass}
T\epsilon\underline{\kappa}\Big(1 - \frac{1}{A}- \sqrt{\frac{T}{4A}\frac{\epsilon^2}{1-4\epsilon^2}}\Big).\nonumber
\end{align}
The last inequality holds by using $\log(x)\leq x-1$. Choosing $\epsilon=\frac{1}{4}\min\Big\{\sqrt{\frac{T}{A}},1\Big\}$ completes the proof.
\end{proof}

\subsection{Regret Lower Bound for Risk-sensitive MDP}
We are now ready to extend the result for RS-MAB and show the regret lower bound for the RS-MDP setting.
\begin{theorem}
\label{thm:RSMDP_lower_bound}
Under Assumption \ref{assumption_derivative_LB}, for any $T>A$, we have
\begin{align}
\inf_{\algr\in\algrandclass}\sup_{\mathbb{M}\in\classRSMDP}\E_{\algd\sim\algr}\E_{\mathbb{M}}^{\algd}\big[{\rm regret}(T)\big]=\Omega\Big(\sqrt{HSAT\underline{\kappa}^2}\Big),   
\end{align}
where $\E_{\mathbb{M}}^{\algd}$ denotes the expectation following algorithm $\algd$ under the RS-MDP $\mathbb{M}$, and ${\rm regret}(T)$ denotes the regret for a total time $T$. 
\end{theorem}
{\color{black}Theorem \ref{thm:RSMDP_lower_bound} establishes an instance-independent lower bound for RS-MDPs. Compared with Theorem \ref{thm:VIUCB_regret}, Algorithm \ref{alg:VICNUCB} matches the lower bound in terms of $A$ and $T$, with only a small polynomial gap in $H$, $S$, the Lipschitz constant of the utility function $\kappa$, and a logarithmic factor of the model parameters. The source of this gap arises from the same trade-off discussed in Section \ref{sec:4.3}.}


\begin{proof}[Proof of Theorem \ref{thm:RSMDP_lower_bound}] Given the results in Theorem  \ref{thm:RSMAB_lower_bound}, the proof for RS-MDP follows directly  from \citet{auer2008near}.
We first construct a special MDP with 3 states $\calS=\{s^0,s^1,s^2\}$. For any action $a\in\calA$, $\prob^S(s^1|s^0,a) = 1/2$ for $a \neq a^i$, and  $\prob^S(s^1|s^0,a) = 1/2+\epsilon$ if $a = a^i$. $\prob^S(s^1|s^1,a)=1$ for any $a\in\calA$, and $\prob^S(s^2|s^2,a)=1$ for any $a\in\calA$. $R(s^1,a)=1$ for any $a\in\calA$, and the reward is 0 from $s^0$ or $s^2$.

Then from the initial state $s^0$, the problem becomes a multi-armed bandit problem. The regret lower bound is $\Omega\Big(\sqrt{HAT\underline{\kappa}^2}\Big)$.

Next, make $S/3$ copies of these MDPs with independent best-action index $I\sim{\rm unif\{1,\cdots,A\}}$. Initializing at each copy for ${3K}/S$ times, and the regret is lower-bounded by
\begin{align*}
\Omega\Big(\frac{S}{3}\sqrt{3HAT\underline{\kappa}^2/S}\Big)=\Omega\Big(\sqrt{HSAT\underline{\kappa}^2}\Big).
\end{align*}
\end{proof}
}

\section{Conclusion}
In conclusion, to understand MDP with general utility functions, we develop two efficient RL algorithms. Specifically, we propose the VICN Algorithm when a simulator is accessible and prove the corresponding sample complexity of order $\Tilde{\calO}\left(\frac{H^7SA\kappa^2 }{\epsilon^2}\right)$.  Without access to the simulator, we propose the VICN-UCB Algorithm and prove that its regret upper bound is of order {$\Tilde{\calO}\left(\sqrt{H^4S^2AT\kappa(\lambda+\eta)}+\sqrt{H^3SAT\kappa^2{\iota_2}}\right)$}, {\color{black} which matches our proposed instance-independent regret lower bound, with order $\Omega\Big(\sqrt{HSAT\underline{\kappa}^2}\Big)$, in $A$ and $T$.} 
Another beneficial direction to explore is the development of an efficient deep-learning-based algorithm under general utility for large-scale applications. This is crucial because, despite extensive research on deep RL algorithms in the conventional risk-neutral setting \citep{shen2020deep, fan2020theoretical, yin2023sample}, the role of risk sensitivity in deep RL algorithms remains less explored. 





\bibliography{references}  
\appendix
\section{Proofs for Section \ref{sec:enlarged_state_space}}
\subsection{Proof of Theorem \ref{thm:markovian_policy}}
\label{sec:append_A1}

For any $h\in[H]$ and $\psi\in\Psi$, define the value function under a given history $g_h\in G_h$ as:
\begin{align}
J_h^{\psi}(g_h):=\E^{\pi}\left.\left[
U\Big(\sum_{h'=1}^H r_{h'}\Big)\, \right|\, g_h
\right].
\end{align}
Moreover, define 
\begin{align}
J_{H+1}^{\psi}(g_{H+1}):=U\left(\sum_{h'=1}^H r_{h'}\right).
\end{align}

\begin{lemma}[Bellman Equation for History-dependent Policy]
\label{lemma_bellman_equation_history_dependent}
For any $h\in[H]$ and $\psi\in\Psi$, we have 
\begin{align}
\label{equation_bellman_equation_history_dependent_1}
J^\psi_h(g_h)=\E^\psi[J^\psi_{h+1}(g_{h+1})  \mid g_h].
\end{align}
\end{lemma}
\begin{proof}[Proof of Lemma \ref{lemma_bellman_equation_history_dependent}]
For any $h\in[H]$, $g_h\in G_h$ and $\psi \in \Psi$, we have
\begin{align}
\E^\psi[J^\psi_{h+1}(g_{h+1}) \mid g_h]&=
\E^\psi\left[\E^\psi\left[U\left(\sum_{h'=1}^{H}r_{h'}\right) \Big| g_{h+1}\right]\, \Big|\, g_h\right]
\label{equation_bellman_equation_history_dependent_2} \\
&=\E^\psi\left.\left[U(\sum_{h'=1}^{H}r_{h'}) \right| g_h\right]
\label{equation_bellman_equation_history_dependent_3}\\
&=J^\psi_h(g_h).
\end{align}
\eqref{equation_bellman_equation_history_dependent_2} holds by the definition of $J_{h+1}^\psi$, and \eqref{equation_bellman_equation_history_dependent_3} holds by the tower property. 
\end{proof}

\begin{lemma}[Optimal Bellman Equation for History-dependent Policies]
\label{lemma_optimal_bellman_equation_history_dependent}
Define the optimal history-dependent policy as:
\begin{align}
\psi_h^*(g_h) \in\argmax_{\psi\in\Psi} J_h^{\psi}(g_h).\label{equation_optimal_bellman_equation_history_dependent_1}
\end{align}
Then we have the optimal Bellman equation for history-dependent policies:
\begin{align}\label{eq:Jpsistar}
J_h^{\psi^*}(g_h)=\max_{a\in\calA}\E\left.\left[J_{h+1}^{\psi^*}(g_{h+1}) \,\right| \,g_h,a_h=a\right].
\end{align}
\end{lemma}

\begin{proof}[Proof of Lemma \ref{lemma_optimal_bellman_equation_history_dependent}]
For any $\psi' \in \Psi$, let $a':=\psi'_h(g_h)$. Then we have
\begin{align}
\max_{a\in\calA}\E[J_{h+1}^{\psi^*}(g_{h+1}) \mid g_h,a_h = a]&\geq
\E[J_{h+1}^{\psi^*}(g_{h+1})\mid g_h,a_h = a']\nonumber\\
&\geq \E[J_{h+1}^{\psi'}(g_{h+1})\mid g_h,a_h = a']\label{equation_optimal_bellman_equation_history_dependent_2}\\
&=J_{h}^{\psi'}(g_{h}).\label{equation_optimal_bellman_equation_history_dependent_3}
\end{align}
\eqref{equation_optimal_bellman_equation_history_dependent_2} holds because $J_h^{\psi^*}$ is the optimal value function, and \eqref{equation_optimal_bellman_equation_history_dependent_3} is by Lemma \ref{lemma_optimal_bellman_equation_history_dependent}. Since \eqref{equation_optimal_bellman_equation_history_dependent_3} holds for any $\psi'\in\Psi$, we have
\begin{align}
\max_{a\in\calA}\E[J_{h+1}^{\psi^*}(g_{h+1}) \mid g_h,a_h = a]
&\geq  
J_{h}^{\psi^*}(g_{h}).\label{equation_optimal_bellman_equation_history_dependent_4}
\end{align}
Next, let $$a^*\in\argmax_{a\in\calA}\E[J_{h+1}^{\psi^*}(g_{h+1}) \mid g_h,a],$$
which leads to a policy that is history-dependent. Therefore, 
\begin{align}
\max_{a\in\calA}\E[J_{h+1}^{\psi^*}(g_{h+1}) \mid g_h,a_h = a]\leq    \max_{\psi'\in\Psi}J_{h}^{\psi'}(g_{h})\label{equation_optimal_bellman_equation_history_dependent_5}.
\end{align}
Combining \eqref{equation_optimal_bellman_equation_history_dependent_4} and \eqref{equation_optimal_bellman_equation_history_dependent_5} yields the desired result.
\end{proof}

\begin{proof}[Proof of Theorem \ref{thm:markovian_policy}]
First, we prove part (a). For any $\pi\in\Pi$, at any timestamp $h\in [H]$, at any enlarged state $(s,y)\in\calS\times\calY_h$, we have 
\begin{align}
&\E^\pi\Big[V^{\pi}_{h+1}(s^\prime,y+r_h)\Big| s_h=s,y_h=y \Big]\nonumber\\
&=\E^{\pi}\Big[
\E^{\pi}\big[U\big(\sum_{h^\prime=h+1}^H r_{h^\prime}+y+r_h\big)\big| \,s_{h+1}=s',y_{h+1}=y+r_h\big]\Big| s_h=s,y_h=y\Big]\label{equation_optimal_markovian_policy_1}\\
&=\E^{\pi}\Big[
U\Big(\sum_{h^\prime=h}^H r_{h^\prime}+y\Big)\Big| 
s_h=s,y_h=y \Big]\label{equation_optimal_markovian_policy_2}\\
&=V_h^\pi(s,y).\label{equation_optimal_markovian_policy_3}
\end{align}
\eqref{equation_optimal_markovian_policy_1} holds by the definition of $V_{h+1}^\pi$, \eqref{equation_optimal_markovian_policy_2} holds by tower property, and \eqref{equation_optimal_markovian_policy_3} holds by the definition of $V_h^\pi$.

Next, part (b) is a direct consequence of Lemma \ref{lemma_optimal_bellman_equation_history_dependent}.

Finally, we prove part (c) by induction. 
Suppose $J_{h'}^{\psi^*}(g_{h'})=V_{h'}^{\pi^*}(s_{h'},y_{h'})$ holds for $h'=h+1,h+2,\dots,H+1$. At timestamp $h$, we have
\begin{align}
J_h^{\psi^*}(g_h)&=\max_{a\in\calA}\E[{J_{h+1}^{\psi^*}(g_{h+1})}\mid g_h,a_h = a]\label{equation_optimal_markovian_policy_4}\\
&=\max_{a\in\calA}\E[V_{h+1}^{\pi^*}(s_{h+1},y_{h+1})\mid g_h,a_h=a]\label{equation_optimal_markovian_policy_5}\\
&=V_{h}^{\pi^*}(s_h,y_h).\label{equation_optimal_markovian_policy_6}
\end{align}
Here \eqref{equation_optimal_markovian_policy_4} holds by the optimal Bellman equation for history-dependent policy, \eqref{equation_optimal_markovian_policy_5} holds by induction, and  \eqref{equation_optimal_markovian_policy_6} holds by the optimal Bellman equation for Markovian policy. 
Finally, we can check that at the last step, 
\begin{align}
J_{H+1}^{\psi^*}(g_{H+1})=V_{H+1}^{\pi^*}(s_{H+1},y_{H+1})=U(y_{H+1}).\nonumber
\end{align}
This concludes that $J_{h}^{\psi^*}(g_h)=V_{h}^{\pi^*}(s_h,y_h)$ for all $h\in[H+1]$. Given the optimal Markovian policy $\pi^*$, the converted history-dependent policy from \eqref{equ:policy_tranlation} will take the same action as $\pi^*$ and is therefore optimal. This finishes the proof.
\end{proof}

\subsection{Proof of Theorem \ref{thm:Lipschitz}}
\label{sec:append_A2}
\begin{proof}[Proof of Theorem \ref{thm:Lipschitz}]
We prove the statement by backward induction. Suppose  $V_{h'}^{{\pi^*}}(s,\cdot)$ is $\kappa$-Lipschitz $\forall s \in \calS$ and $h'=h+1,h+2\dots H+1$. We first show that for fixed $s$ and $a$, $ Q_h^{{\pi^*}}(s,\cdot,a)$ is also $\kappa$-Lipschitz. For any $y_1,y_2\in{\calY_h}$,
\begin{align}
& \left|Q_h^{{\pi^*}}(s,y_1,a)-Q_h^{{\pi^*}}(s,y_2,a) \right|\nonumber \\
&= \left|\E_{s^\prime\sim\prob_h^S(\cdot\mid s,a),r\sim R_h(\cdot\mid s,a)}\left[V_{h+1}^{{\pi^*}}(s^\prime,y_1+r)\right] -\E_{s^\prime\sim\prob_h^S(\cdot\mid s,a),r\sim R_h(\cdot\mid s,a)}\left[V_{h+1}^{{\pi^*}}(s^\prime,y_2+r)\right]\right| \label{thm_Lipschitz_equation_1}\\
&=\left|\E_{s^{\prime}\sim \prob^S_h(\cdot\mid s,a), r \sim R_h(\cdot\mid s,a)} \left[V_{h+1}^{{\pi^*}}(s^\prime,y_1+r)-V_{h+1}^{{\pi^*}}(s^\prime,y_2+r)\right] \right| \nonumber\\
&\leq\E_{s^{\prime}\sim \prob^S_h(\cdot\mid s,a), r \sim R_h(\cdot\mid s,a)} \left[\left|[V_{h+1}^{{\pi^*}}(s^\prime,y_1+r)-V_{h+1}^{{\pi^*}}(s^\prime,y_2+r)\right|\right] \nonumber\\
&\leq\E_{s^{\prime}\sim \prob^S_h(\cdot\mid s,a), r \sim R_h(\cdot\mid s,a)} \left[\kappa |y_1-y_2|\right] \label{thm_Lipschitz_equation_2}\\
&= \kappa |y_1-y_2|.\nonumber
\end{align}
\eqref{thm_Lipschitz_equation_1} holds by applying the Bellman equation from Theorem \ref{thm:markovian_policy}. \eqref{thm_Lipschitz_equation_2} holds by induction.
Note that $\pi^*$ is an optimal policy, so we have $\pi^*(s,y)\in \text{arg}\max_{a\in \calA} Q_h^{{\pi^*}}(s,y,a)$. Therefore,
\begin{align}
\left|V_h^{{\pi^*}}(s,y_1)-V_h^{{\pi^*}}(s,y_2) \right|
&= \left|Q_h^{{\pi^*}}(s,y_1,{\pi^*}(s,y_1))-Q_h^{{\pi^*}}(s,y_2,{\pi^*}(s,y_2))\right|\nonumber\\
&=\left|\max_{a\in \calA} Q_h^{{\pi^*}}(s,y_1,a)-\max_{a\in \calA} Q_h^{{\pi^*}}(s,y_2,a) \right|\label{thm_Lipschitz_equation_3}\\
&\leq \max_{a\in \calA}\left |Q_h^{{\pi^*}}(s,y_1,a)-Q_h^{{\pi^*}}(s,y_2,a)\right|\nonumber\\
&\leq \kappa |y_1-y_2|.\nonumber
\end{align}
\eqref{thm_Lipschitz_equation_3} holds by applying the optimal Bellman operator thanks to Theorem \ref{thm:markovian_policy}. This implies that $ V_h^{{\pi^*}}(s,\cdot)$ is $\kappa$-Lipschitz in $y$. Lastly, we can check at the final timestamp $h=H+1$, $V_{H+1}^{{\pi^*}}(s,y)=U(y)$ is $\kappa$-Lipschitz. This finishes the proof.
\end{proof}

\section{Proofs for Section \ref{sec:discretized_env}}
\label{sec:append_B}
\begin{proof}[Proof of Proposition \ref{lemma:v_difference_ERSMDP_BARMDP}]
First, for any $h\in[H]$, $(s,y,a)\in\calS\times\barY_h\times\calA$ and $\pi\in\barPi$, we have the following equations hold:
\begin{align}
\barQ^{\pi}_h(s,y,a)&=\E_{s^\prime\sim\prob_h^S(\cdot\mid s,a)}\left[
\sum_{r'\in\barI} \barV_{h+1}^\pi(s^\prime,y+r')\barR_h(r'\mid s,a)
\right]\nonumber\\
&=\E_{s^\prime\sim\prob_h^S(\cdot\mid s,a)}\left[
\sum_{r'\in\barI} \barV_{h+1}^\pi(s^\prime,y+r')\int_{r'-\frac{\epsilon_o}{2}}^{r'+\frac{\epsilon_o}{2}}R_h({\rm d}r\mid s,a)
\right]\label{equ:lemma_v_difference_ERSMDP_BARMDP_4}\\
&=\E_{s^\prime\sim\prob_h^S(\cdot\mid s,a)}\left[
\sum_{r'\in\barI} \int_{r'-\frac{\epsilon_o}{2}}^{r'+\frac{\epsilon_o}{2}}\barV_{h+1}^\pi(s^\prime,y+r')R_h({\rm d}r\mid s,a)
\right]\label{equ:lemma_v_difference_ERSMDP_BARMDP_5}\\
&=\E_{s^\prime\sim\prob_h^S(\cdot\mid s,a)}\left[
\int_{0}^{1}\barV_{h+1}^\pi(s^\prime,y+\projR{r})R_h({\rm d}r\mid s,a)
\right].\label{equ:lemma_v_difference_ERSMDP_BARMDP_6}
\end{align}
\eqref{equ:lemma_v_difference_ERSMDP_BARMDP_4} holds by the definition of $\barR_h$. \eqref{equ:lemma_v_difference_ERSMDP_BARMDP_5}-\eqref{equ:lemma_v_difference_ERSMDP_BARMDP_6} holds by straightforward calculations. 

Now we are ready to prove Proposition \ref{lemma:v_difference_ERSMDP_BARMDP} by induction. Suppose \eqref{equ:lemma_v_difference_ERSMDP_BARMDP_1} holds for $h' = h+1,h+2\cdots H$. Then for $h'=h$ we have for any $s\in \calS$ and $y\in\barY$, 
\begin{align}
\left|V_h^{\pi^*}(s,y)-\barV^{{\barpi^*}}_h(s,y)\right|&=\left|\max_{a\in\calA}Q_h^{\pi^*}(s,y,a)-\max_{a\in\calA}\barQ^{{\barpi^*}}_h(s,y,a)\right|\label{equ:lemma_v_difference_ERSMDP_BARMDP_2}\\
&\leq \max_{a\in\calA}\left|Q_h^{\pi^*}(s,y,a)-\barQ^{{\barpi^*}}_h(s,y,a)\right|.\label{equ:lemma_v_difference_ERSMDP_BARMDP_3}
\end{align}
\eqref{equ:lemma_v_difference_ERSMDP_BARMDP_2} holds because $\pi^*$ and ${\barpi^*}$ are the optimal policies for $\RSMDP$ and $\BARMDP$ respectively. \eqref{equ:lemma_v_difference_ERSMDP_BARMDP_3} used the fact that {the} $\max$-operator is a contraction mapping. Now to further bound \eqref{equ:lemma_v_difference_ERSMDP_BARMDP_3}, observe that by {the} Bellman equation,
\begin{equation}
\label{equ:lemma_v_difference_ERSMDP_BARMDP_3.5}
Q_h^{\pi^*}(s,y,a)=\E_{s^\prime\sim\prob_h^S(\cdot\mid s,a)} \left[ 
\int_{0}^1 
V_{h+1}^{\pi^*}(s^\prime,y+r)R_h({\rm d}r\mid s,a)\right].
\end{equation}
Combining  \eqref{equ:lemma_v_difference_ERSMDP_BARMDP_6},  \eqref{equ:lemma_v_difference_ERSMDP_BARMDP_3} and \eqref{equ:lemma_v_difference_ERSMDP_BARMDP_3.5} gives
\begin{align} 
&\left|V_h^{\pi^*}(s,y)-\barV^{{\barpi^*}}_h(s,y)\right|\nonumber\\
&\leq \max_{a\in\calA}\left|\E_{s^\prime\sim\prob_h^S(\cdot\mid s,a)} \left[ 
\int_{0}^1 \left(V_{h+1}^{\pi^*}(s^\prime,y+r)-\barV_{h+1}^{{\barpi^*}}(s^\prime,y+\projR{r})\right)R_h({\rm d}r\mid s,a)
\right] \right|\nonumber\\
&\leq \max_{a\in\calA}\E_{s^\prime\sim\prob_h^S(\cdot\mid s,a)} \left[ 
\int_{0}^1 \left|V_{h+1}^{\pi^*}(s^\prime,y+r)-\barV_{h+1}^{{\barpi^*}}(s^\prime,y+\projR{r})\right|R_h({\rm d}r\mid s,a)
\right] 
\label{equ:lemma_v_difference_ERSMDP_BARMDP_7}.
\end{align}
Now it suffices to bound the integral: 
\begin{eqnarray}
&&\int_{0}^1 \left|V_{h+1}^{\pi^*}(s^\prime,y+r)-\barV_{h+1}^{{\barpi^*}}(s^\prime,y+\projR{r})\right|R_h({\rm d}r\mid s,a)\nonumber\\
&\leq&\int_{0}^1 \left|V_{h+1}^{\pi^*}(s^\prime,y+r)-
V_{h+1}^{\pi^*}(s^\prime,y+\projR{r})\right|R_h({\rm d}r\mid s,a)\nonumber\\
&&\quad\quad+\int_{0}^1 \left|V_{h+1}^{\pi^*}(s^\prime,y+\projR{r})-\barV_{h+1}^{{\barpi^*}}(s^\prime,y+\projR{r})\right|R_h({\rm d}r\mid s,a)\label{equ:lemma_v_difference_ERSMDP_BARMDP_8}\\
&\leq&\sup_{r'\in[0,1]}\left|V_{h+1}^{\pi^*}(s^\prime,y+r')-
V_{h+1}^{\pi^*}(s^\prime,y+\projR{r'})\right|\left(\int_{0}^1 R_h({\rm d}r\mid s,a)\right)\nonumber\\
&&\quad\quad+\sup_{r'\in[0,1]} \left|V_{h+1}^{\pi^*}(s^\prime,y+\projR{r'})-\barV_{h+1}^{{\barpi^*}}(s^\prime,y+\projR{r'})\right|\left(\int_{0}^1 R_h({\rm d}r\mid s,a)\right)\label{equ:lemma_v_difference_ERSMDP_BARMDP_8.5}\\
&\leq& \kappa\epsilon_o+ (H-h)\kappa \epsilon_o= (H-h+1)\kappa \epsilon_o.\label{equ:lemma_v_difference_ERSMDP_BARMDP_9}
\end{eqnarray}
Here \eqref{equ:lemma_v_difference_ERSMDP_BARMDP_8.5} holds by applying H\"older's inequality. To prove \eqref{equ:lemma_v_difference_ERSMDP_BARMDP_9}, note that 
in \eqref{equ:lemma_v_difference_ERSMDP_BARMDP_8.5}, 
$$\left|V_{h+1}^{\pi^*}(s^\prime,y+r')-
V_{h+1}^{\pi^*}(s^\prime,y+\projR{r'})\right|\leq \kappa\epsilon_o$$ holds due to Theorem \ref{thm:Lipschitz}, and 
$$\left|V_{h+1}^{\pi^*}(s^\prime,y+\projR{r'})-\barV_{h+1}^{{\barpi^*}}(s^\prime,y+\projR{r'})\right|\leq(H-h)\kappa \epsilon_o$$ holds by induction. Combine \eqref{equ:lemma_v_difference_ERSMDP_BARMDP_9} with \eqref{equ:lemma_v_difference_ERSMDP_BARMDP_7},  we get 
\begin{align*}
\left|V_h^{\pi^*}(s,y)-\barV^{{\barpi^*}}_h(s,y)\right|\leq (H-h+1) \kappa\epsilon_o.
\end{align*}
Finally, we can check at the last timestamp $h'=H$, 
\begin{align}
\left|V_H^{\pi^*}(s,y)-\barV_H^{{\barpi^*}}(s,y)\right|&=\left|\int_{0}^1 \left(U(y+r)-U(y+\projR{r})\right)R_H({\rm d}r\mid s,a)\right|\label{equ:lemma_v_difference_ERSMDP_BARMDP_10}\\
& \leq \left|\int_{0}^1 \kappa \epsilon_o R_H({\rm d}r\mid s,a)\right|\nonumber\\
&=(H-H+1) \kappa \epsilon_o.\nonumber
\end{align}
\eqref{equ:lemma_v_difference_ERSMDP_BARMDP_10} holds by applying \eqref{equ:lemma_v_difference_ERSMDP_BARMDP_6} and by the fact that $V_{H+1}^{\pi^*}(s,y)=\barV_{H+1}^{\barpi^*}(s,y)=U(y)$. This finishes the proof.
\end{proof}

\begin{proof}[Proof of Proposition \ref{lemma:Lipschitz_Vbar_pihat}] Let $\bary_1:=\projY{y_1}$ and  $\bary_2:=\projY{y_2}$ for the ease of notation. 
For any $s\in \calS$, $y_1,y_2\in{\calY_h}$,  at any timestamp $h\in[H]$ we have: 
\begin{align}
&\Big|\barV^\barpi_h(s,y_1)-\barV^\barpi_h(s,y_2)\Big|=\Big|\barV^\barpi_h(s,\bary_1)-\barV^\barpi_h(s,\bary_2)\Big|\nonumber\\
&=\Big|\barV^\barpi_h(s,\bary_1)-\barV^{\barpi^*}_h(s,\bary_1)
+\barV^{\barpi^*}_h(s,\bary_1)-\barV^{\barpi^*}_h(s,\bary_2)
+\barV^{\barpi^*}_h(s,\bary_2)-\barV^\barpi_h(s,\bary_2)\Big|. \label{equ:lemma_Lipschitz_Vbar_pihat_2}
\end{align}
By a similar version to Theorem \ref{thm:Lipschitz}, 
\begin{equation}\label{equ:lemma_Lipschitz_Vbar_pihat_5}
\left|\barV^{\barpi^*}_h(s,\bary_1)-\barV^{\barpi^*}_h(s,\bary_2)\right|\leq\kappa\left|\bary_1-\bary_2\right|.
\end{equation}
Combining \eqref{equ:lemma_Lipschitz_Vbar_pihat_0}, \eqref{equ:lemma_Lipschitz_Vbar_pihat_2},  and \eqref{equ:lemma_Lipschitz_Vbar_pihat_5} yields the desired result.
\end{proof} 

\begin{proof}[Proof of Proposition \ref{lemma:V_Vbar_difference}]
We prove the statement by induction.
Suppose \eqref{equ:lemma_V_Vbar_difference_1} holds for $h'=h+1,h+2\cdots H+1$. For the ease of notation,  let $\bary:=\projY{y}$ for any $y\in{\calY_h}$ and  $\barr:=\projR{r}$ for $r\in[0,1]$. Then we have for any $\barpi\in\barPi$, for any $(s,y)\in\calS\times{\calY_h}$, 
\begin{align}
&\left|V_h^{\Gamma(\barpi)}(s,y)-\barV_h^\barpi(s,\bary)\right|\nonumber\\
&=\left|Q_h^{\Gamma(\barpi)}(s,y,\barpi_h(s,\bary))-\barQ_h^\barpi(s,\bary,\barpi_h(s,\bary))\right|\label{equ:lemma_V_Vbar_difference_2}\\
&\leq\E_{s^\prime\sim\prob^S_h(\cdot\mid s,\barpi_h(s,\bary))}\left[\int_{0}^1\left|
V_{h+1}^{\Gamma(\barpi)}(s^\prime,y+r)-\barV_{h+1}^\barpi(s^\prime,\bary+\barr)\right|R_h({\rm d}r\mid s,\barpi_h(s,\bary)) 
\right]\label{equ:lemma_V_Vbar_difference_3}\\
&\leq \E_{s^\prime\sim\prob^S_h(\cdot\mid s,\barpi_h(s,\bary))}\left[\int_{0}^1\left|
V_{h+1}^{\Gamma(\barpi)}(s^\prime,y+r)-\barV_{h+1}^\barpi(s^\prime,y+r)\right|R_h({\rm d}r\mid s,\barpi_h(s,\bary))
\right]\nonumber\\
&\qquad +\E_{s^\prime\sim\prob^S_h(\cdot\mid s,\barpi_h(s,\bary))}\left[\int_{0}^1\left|
\barV_{h+1}^\barpi(s^\prime,y+r)-\barV_{h+1}^\barpi(s^\prime,\bary+\barr)\right|R_h({\rm d}r\mid s,\barpi_h(s,\bary))
\right]\label{equ:lemma_V_Vbar_difference_4}\\
&\leq \E_{s^\prime\sim\prob^S_h(\cdot\mid s,\barpi_h(s,\bary))}\left[\sup_{r\in[0,1]}\left|
V_{h+1}^{\Gamma(\barpi)}(s^\prime,y+r)-\barV_{h+1}^\barpi(s^\prime,y+r)\right|\int_{0}^1R_h({\rm d}r'\mid s,\barpi_h(s,\bary))
\right]\nonumber\\
&\qquad +\E_{s^\prime\sim\prob^S_h(\cdot\mid s,\barpi_h(s,\bary))}\left[\sup_{r\in[0,1]}\left|
\barV_{h+1}^\barpi(s^\prime,y+r)-\barV_{h+1}^\barpi(s^\prime,\bary+\barr)\right|\int_{0}^1R_h({\rm d}r'\mid s,\barpi_h(s,\bary))
\right]. \label{equ:lemma_V_Vbar_difference_5}
\end{align}
\eqref{equ:lemma_V_Vbar_difference_2} holds by the definition of value function, and \eqref{equ:lemma_V_Vbar_difference_3} holds by {the} Bellman equation and by \eqref{equ:lemma_v_difference_ERSMDP_BARMDP_6}. In addition, \eqref{equ:lemma_V_Vbar_difference_4} holds by telescoping sums and \eqref{equ:lemma_V_Vbar_difference_5} holds by H\"older's inequality.
In \eqref{equ:lemma_V_Vbar_difference_5}, 
$$\left| V_{h+1}^{\Gamma(\barpi)}(s^\prime,y+r)-\barV_{h+1}^\barpi(s^\prime,y+r)\right|\leq 2\sum_{h'=h+1}^H \Delta_{h'} + (H-h)\kappa\epsilon_o$$ holds by induction.
By Proposition \ref{lemma:Lipschitz_Vbar_pihat}, 
$$\left|
\barV_{h+1}^\barpi(s^\prime,y+r)-\barV_{h+1}^\barpi(s^\prime,\bary+\barr)\right|\leq 2\Delta_h+\kappa\epsilon_o.$$ 
Combining the above inequality with \eqref{equ:lemma_V_Vbar_difference_5} gives
\begin{align*}
&\left|V_h^{\Gamma(\barpi)}(s,y)-\barV_h^\barpi(s,y)\right|\nonumber\leq 2\sum_{h'=h}^H \Delta_{h'} + (H-h+1)\kappa\epsilon_o.
\end{align*}
Lastly, we can check at the last timestamp $H+1$, 
\begin{align*}
\left|V_{H+1}^{\Gamma(\barpi)}(s,y)-\barV_{H+1}^\barpi(s,y)\right|=\left|U(y)-U(\bary)\right|\leq\kappa\epsilon_o.
\end{align*}
This finishes the proof.
\end{proof}

\section{Deleted contents}
We also have the following observation for the optimal value function.

\begin{proposition}[Convexity of the Optimal Value Function]
\label{proposition:convex}
Given a $\RSMDP$, if $U(\cdot)$ is 
convex, then for any $h\in[H+1]$, the value function of the optimal policy at a given state, $V_h^{\pi^*}(s,\cdot)$,  is also convex in $y$.
\end{proposition}

\begin{proof} [Proof of Proposition \ref{proposition:convex}]
Suppose for any $s\in\calS$, $Q_{h'}^{\pi^*}(s,y,a)$ and $V_{h'}^{\pi^*}(s,y)$ are convex in $y$ for $h'=h+1,\cdots H+1$. Then at timestamp $h$, for any $y_1,y_2\in {\calY_h}$, $\alpha \in (0,1)$, 
\begin{align}
&\alpha Q_h^{\pi^*}(s,y_1,a)+(1-\alpha) Q^{\pi^*}_h(s,y_2,a)\nonumber\\
&=\E_{s^\prime\sim\prob_h^S(\cdot\mid s,a),r\sim R_h(\cdot\mid s,a)}\left[\alpha V^{\pi^*}_{h+1}\left(s^\prime,r+y_1\right)+(1-\alpha)V^{\pi^*}_{h+1}\left(s^\prime,r+y_2\right)\right]\label{proposition_convex_equation_2}\\
&\geq \E_{s^\prime\sim\prob_h^S(\cdot\mid s,a),r\sim R_h(\cdot\mid s,a)}\left[V^{\pi^*}_{h+1}\left(s^\prime,\alpha (r+y_1)+(1-\alpha)(r+y_2)\right)\right]\label{proposition_convex_equation_3}\\
&=\E_{s^\prime\sim\prob_h^S(\cdot\mid s,a),r\sim R_h(\cdot\mid s,a)}\left[V^{\pi^*}_{h+1}\left(s^\prime,r+\alpha y_1+(1-\alpha)y_2\right)\right]\nonumber\\
&=Q^{\pi^*}_h\left(s,\alpha y_1+(1-\alpha)y_2\right),
\label{proposition_convex_equation_4}
\end{align}
where \eqref{proposition_convex_equation_2} holds by the Bellman equation. \eqref{proposition_convex_equation_3} holds by induction, and \eqref{proposition_convex_equation_4} holds again by the Bellman equation. Next, we prove $V_h^{\pi^*}(s,\cdot)$ is convex:
\begin{align}
&\alpha V_h^{\pi^*}(s,y_1)+(1-\alpha) V^{\pi^*}_h(s,y_2)\nonumber\\
&=\max_{a\in\calA} \alpha Q_h^{\pi^*}(s,y_1,a)+ \max_{a\in\calA}(1-\alpha)Q^{\pi^*}_h(s,y_2,a)\nonumber\\
&\geq \max_{a\in\calA}\left( \alpha Q_h^{\pi^*}(s,y_1,a)+ (1-\alpha)Q^{\pi^*}_h(s,y_2,a)\right) \label{proposition_convex_equation_5}\\
&\geq \max_{a\in\calA}\left(  Q_h^{\pi^*}(s,\alpha y_1+(1-\alpha)y_2,a)\right)\label{proposition_convex_equation_6}\\
&=V_h^{\pi^*}(s,\alpha y_1+(1-\alpha)y_2)\nonumber
\end{align}
\eqref{proposition_convex_equation_5} holds by the contraction property of the $\max$-operator, and \eqref{proposition_convex_equation_6} holds by \eqref{proposition_convex_equation_4}. Finally, we can check at timestamp $H+1$, $Q_{H+1}^{\pi^*}(s,y,a)=V_{H+1}^{\pi^*}(s,y)=U(y)$ is convex. This finishes the induction. 
\end{proof}

\end{document}